%% file: draft.tex
\pdfoutput=1

\documentclass[10pt,twocolumn,twoside,journal]{IEEEtran}

\usepackage{cs}
\usepackage{mathsymb}
\usepackage{verbatim}
\usepackage{subfigure}
\usepackage{url}
\usepackage{graphicx}
\graphicspath{{./}{./Fig/}{./Figs/}{./Fig/eps/}{./Fig/pdf/}}

\usepackage{cite}

\usepackage[]{algorithm2e}

\usepackage{amsmath}
\usepackage{url,eucal}
\usepackage{multirow}
\usepackage{pifont}

\def\hset{\mathcal{H}}
\def\Real{\mathbb{R}}
\def\Integer{\mathbb{Z}^+}

\def\brho{{\boldsymbol \rho}}

\def\bHi{{\bh_{i:}^\T}}
\def\bwyi{{\bw_{:y_{i}}}}
\def\bwr{{\bw_{:r}}}

\def\yi{{y_i}}

\def\bxi{{\bx_i}}

\def\Delta{ \pmb{\delta} }

\def\MultiEXP{{\rm MCBoost$^{\rm exp}_{\rm sw}$}\xspace}
\def\MultiLOG{{\rm MCBoost$^{\rm log}_{\rm sw}$}\xspace}
\def\MultiLOGshort{{\rm MCBoost-${\rm log}$}\xspace}
\def\MultiEXPshort{{\rm MCBoost-${\rm exp}$}\xspace}
\def\CVX{\texttt{CVX}\xspace}
\def\paragraph{\textbf}
\def\Lag{{\varLambda}}
\def\rhoir{\rho_{i,r}}
\def\uir{u_{ir}}

\newcommand{\revised}[1]{{#1}}
\newcommand{\secondrev}[1]{{#1}}

\def\cone{{\ding{172}}}
\def\ctwo{{\ding{173}}}
\def\cthree{{\ding{174}}}
\def\cfour{{\ding{175}}}
\def\cfive{{\ding{176}}}
\def\csix{{\ding{177}}}

\def\paragraph{\textbf}
\newcommand{\bigO}{\ensuremath{\mathcal{O}}}

\begin{document}

\title{A scalable stage-wise approach to large-margin multi-class loss based boosting}

\author{
         Sakrapee Paisitkriangkrai,
         Chunhua Shen,
         Anton van den Hengel
\thanks
{
}
\thanks
{
The authors are with Australian Center for Visual Technologies,
and School of Computer Science,
The University of Adelaide, SA 5005, Australia
(e-mail: \{paul.paisitkriangkrai, chunhua.shen, anton.vandenhengel\}@adelaide.edu.au).
Correspondence should be addressed to C. Shen.
}
\thanks
 {
 This work was in part supported by Australian Research Council Future
 Fellowship FT120100969.
 }
}

\markboth{Manuscript}
{Paisitkriangkrai
\MakeLowercase{\textit{et al.}}: A scalable approach to large-margin multi-class boosting}

\maketitle

\begin{abstract}

     We present a scalable and effective classification model
     to train multi-class boosting for multi-class classification
     problems.
    Shen and Hao introduced a direct formulation of multi-class boosting in the
    sense that it directly maximizes the multi-class margin \cite{Shen2011Direct}.
    The major problem of their approach is its high
    computational complexity for training, which hampers its application on
    real-world problems.
    In this work, we propose a scalable and simple stage-wise
    multi-class boosting method, which
    also directly maximizes the multi-class margin.
    Our approach offers a few advantages:
    1)
    it is simple and computationally efficient to train.  The approach
    can speed up the training time by more than two orders of
    magnitude without sacrificing the classification accuracy.
    2)
    Like traditional AdaBoost, it is less sensitive to the
    choice of parameters and empirically demonstrates excellent
    generalization performance.
    Experimental results on challenging multi-class machine
    learning and vision tasks demonstrate that the proposed approach
    substantially improves the convergence rate and accuracy of the
    final visual detector at no additional computational cost
    compared to existing multi-class boosting.

\end{abstract}

\begin{IEEEkeywords}
    Boosting, multi-class classification,
    column generation, convex optimization
\end{IEEEkeywords}

\tableofcontents

\section{Introduction}

Multi-class classification is one of the fundamental problems in
machine learning and computer vision, as many real-world problems
involve predictions which require an instance to be assigned to one of
number of classes.  Well known problems include handwritten character
recognition \cite{Knerr1992Handwritten},
object recognition \cite{Griffin2007Caltech}, and
scene classification \cite{Lazebnik2006Beyond}.
Compared to the well studied binary form of the classification problem,
multi-class problems are considered more difficult to solve,
especially as a number of classes increases.

In recent years a substantial body of work related to multi-class boosting has arisen in the literature.
Many of these works attempt to achieve multi-class boosting by reducing or
reformulating the task into a series of binary boosting problems.
Often this is done through the use of output coding matrices.
The common $1$-vs-all and $1$-vs-$1$ schemes are a particular example of this approach,
in which the coding matrices are predefined.
The drawback of coding-based approaches is that they do not rapidly converge to low training errors on difficult data sets and many weak classifiers need to be learned (as is shown below in our experiments).
As a result these algorithms fail to deliver the level of performance required to process large data sets, or to achieve real-time data processing.

The aim of this paper is to develop a more direct boosting algorithm applicable to multi-class problems that will achieve the effectiveness and efficiency of previously proposed methods for binary classification.
To achieve our goal we exploit the efficiency of the coordinate descent algorithm, \eg, AdaBoost \cite{Schapire1999Boosting}, along with more effective and direct formulations of multi-class boosting known as MultiBoost \cite{Shen2011Direct}.
Our proposed approach is simpler than coding-based multi-class boosting since we do not need to learn output coding matrices.
The approach is also fast to train, less sensitive to the choice of parameters chosen
and has a comparable convergence rate to MultiBoost.
Furthermore, the approach shares a similar property to $\ell_1$-constrained maximum margin classifiers, in that it converges asymptotically to the $\ell_1$-constrained solution.

Our approach is based on a novel stage-wise multi-class form of boosting which bypasses error correcting codes by directly learning base classifiers and weak classifiers' coefficients.
The final decision function is a weighted average of multiple weak classifiers.
The work we present in this paper intersects with several successful practical works, such as multi-class support vector machines \cite{Crammer2002Algorithmic}, AdaBoost \cite{Schapire1999Boosting} and column generation based boosting \cite{Shen2010Boosting}.

Our main contributions are as follows:
\begin{itemize}
    \item  Our approach is the first greedy stage-wise multi-class boosting algorithm which
does not rely on codewords and which directly optimizes the boosting
objective function.
In addition, our approach converges asymptotically to the
$\ell_1$-constrained solution;
\item
    We show that our minimization problem shares a connection with those derived from coordinate descent methods.
In addition, our approach is less prone to over-fitting as techniques,
such as shrinkage, can be easily adopted;
\item
Empirical results demonstrate that the approach exhibits the same
classification performance as the state-of-the-art multi-class
boosting classifier \cite{Shen2011Direct}, but is significantly
faster to train, and {\em orders of magnitude more scalable}.
We have made the source code of the proposed boosting methods
    accessible at:

\url{cs.adelaide.edu.au/users/chhshen/projects/SWMCBoost/}.
\end{itemize}

The remainder of the paper is organized as follows.
Section~\ref{sec:related} reviews related works on multi-class
boosting.
Section~\ref{sec:approach} describes the details of our proposed approach,
including its computational complexity, and discusses various aspects
related to its convergence and generalization performance.
Experimental results on machine learning and computer vision data sets
are presented in Section~\ref{sec:exp}.
Section~\ref{sec:con} concludes the paper  with directions
for possible future work.

\section{Related work}
\label{sec:related}
There exist a variety of  multi-class boosting algorithms in the literature.
Many of them solve multi-class learning tasks by reducing multi-class problems to multiple binary classification problems.
We briefly review some well known boosting algorithms here in order to illustrate the novelty of the proposed approach.%

Coding-based boosting was one of the earliest  multi-class boosting algorithms proposed (see, for example, AdaBoost.MH \cite{Schapire1999Improved}, AdaBoost.MO \cite{Schapire1999Improved}, AdaBoost.ECC \cite{Guruswami1999Multiclass}, AdaBoost.SIP \cite{Zhang2009Finding} and JointBoost \cite{Torralba2007Sharing}).
Coding-based approaches perform multi-class classification by combining the outputs of a set of binary classifiers.  This includes popular methods such as $1$-vs-all and $1$-vs-$1$, for example.
Typically, a coding matrix, $\bM \in \{-1,0,+1\}^{n \times k}$, is constructed (where $n$ is the length of a codeword and $k$ is the number of classes).
The algorithm learns a binary classifier, $\hbar_j(\cdot)$, corresponding to a single column\footnote{Each column of $\bM$ defines a binary partition of $k$ classes over data.} of $\bM$ in a stage-wise manner.
Here $\hbar(\cdot)$ is a function that maps an input $\bx$ to $\{-1, +1\}$.
A test instance is classified as belonging to the class associated with the codeword closest in Hamming distance to the sequence of predictions generated by $\hbar_1(\cdot), \cdots, \hbar_n(\cdot)$.
The final decision function for a test datum $\bx$ is
$
F(\bx) = \argmax_{r=1,\cdots,k} \, \sum_{j=1}^{n}  w_j \hbar_j(\bx) m_{jr},
$
where $\bw$ is the weight vector and the $(j,r)$ entry of $\bM$ is $m_{jr}$.
Clearly, the performance of the algorithm is largely influenced by the quality of the coding matrices.
Finding optimum coding matrices, which means identifying classes which should be grouped together, is often non-trivial.
Several algorithms, \eg, max-cut and random-half, have been proposed to build optimal binary partitions for coding matrices \cite{Schapire1997Using}.
Max-cut finds the binary partitions that maximize the error-correcting ability of coding matrix while random-half randomly splits the classes into two groups.
Li\cite{Li2006Multiclass} points out that random-half usually performs better than max-cut because the binary problems formed by max-cut are usually too hard for base classifiers to learn.
Nonetheless, both max-cut and random-half do not achieve the best performance as they do not consider the ability of base classifiers in the optimization of coding matrices.
In contrast, our proposed approach bypasses the learning of output coding by learning base classifiers and weak classifiers' coefficients directly.

Another related approach, which trains a similar decision function, is the multi-class boosting of Duchi and Singer known as GradBoost \cite{Duchi2009Boosting}.
The main difference between GradBoost and the method we propose here is that GradBoost does not directly optimize the boosting objective function.
GradBoost bounds the original non-smooth $\ell_1$ optimization problem by a quadratic function.
It is not clear how well the surrogate approximates the original objective function.
In contrast, our approach solves the original loss function, which is the approach of AdaBoost and LogitBoost.
Shen and Hao have introduced a direct formulation of multi-class
boosting in the sense that it directly maximizes the multi-class
margin. By deriving a meaningful
Lagrange dual problem, column generation is used to design a fully
corrective boosting method \cite{Shen2011Direct}.
The main issue of \cite{Shen2011Direct} is its extremely heavy
computation burden, which hampers its application on real data sets.
Unlike their work, the proposed approach learns a classification model in a stage-wise manner.
In our work, only the coefficients of the latest weak classifiers need to be updated.
As a result, our approach is significantly more computationally efficient and robust
to the regularization parameter value chosen.
Compared to \cite{Shen2011Direct}, at each boosting
iteration, our approach only needs to solve for $k$ variables instead
of $k \cdot t$ variables, where $k$ is the number of classes and $t$ is the
number of current boosting iterations.
This significant reduction in the size of the problem to be solved at
each iteration is responsible for the {\em orders of magnitude
reduction} in
training time required.

\subsection{Notation}

Bold lower-case letters, \eg, $\bw$, denote column vectors and bold upper-case letters,
\eg, $\bW$, denote matrices.
Given a matrix $\bW$, we write the $i$-th row of $\bW$ as $\bw_{i:}^\T$
and the $j$-th column as $\bw_{:j}$.
The $(i,j)$ entry of $\bW$ is $w_{ij}$.
Let $(\bxi, \yi)_{i=1}^{m}$ be the set of training data,
where $\bxi \in \Real^d$ represents an instance, and $\yi \in \{1,2,\cdots,k\}$
 the corresponding class label (where $m$ is the number of training samples and $k$ is the number of classes).
We denote by $\hset$ a set of all possible outputs of weak classifiers
where the size of $\hset$ can be infinite.
Let $\hbar(\cdot)$ denote a binary weak classifier which projects an instance
$\bx$ to $\{-1,+1\}$.
By assuming that we learn a total of $n$ weak classifiers,
the output of weak learners can be represented as $\bH \in \Real^{m \times n}$, where
$h_{ij}$ is the label predicted by weak classifier $\hbar_j(\cdot)$ on the training data $\bxi$.
Each row $\bh_{i:}^\T$ of the matrix $\bH$ represents the output of all weak classifiers when applied to a single training instance $\bxi$.
We build a classifier of the form,
\begin{align}
    \label{EQ:eqFx}
        F(\bx) = \argmax_{r = 1,\cdots,k} {\textstyle \sum_{j=1}^n }  \hbar_j(\bx) w_{jr},
\end{align}
where $\bW \in \Real^{n \times k}$.
Each column of $\bW$, $\bw_{:r}$,
contains coefficients of the linear classifier for class $r$ and each row of $\bW$,
$\bw_{j:}^\T$, consists of the coefficients for the weak classifier $\hbar_j(\cdot)$ for all
class labels.
The predicted label is the index of the column of $\bW$ attaining the highest sum.

\section{Our approach}
\label{sec:approach}

In order to classify an example $(\bxi, \yi)$ correctly,
$\bHi \bwyi$ must be greater than $\bHi \bwr$, for any $r \neq \yi$.
In this paper, we define a set of margins associated with a training example as,
\begin{align}
    \label{EQ:eqx}
        \rho_r(\bxi, \yi) = \rho_{i,r} = \bHi \bwyi - \bHi \bwr, r = 1, \cdots, k.
\end{align}
The training example $\bxi$ is correctly classified only when $\rho_{i,r} \geq 0$.
In boosting, we train a linear combination of basis functions (weak classifiers) which minimizes a given loss function over predefined training samples.
This is achieved by searching for the dimension which gives the steepest descent in the loss and assigning its coefficient accordingly at each iteration.
Commonly applied loss functions are exponential loss of AdaBoost \cite{Schapire1999Boosting}
and binomial log-likelihood loss of LogitBoost
\cite{Friedman2000Additive}. They are:
\begin{align*}
   & \rm{Exponential} & L_{\rm{exp}}(\bx_i, y_i) &= \exp( - \rho_{i,r}
   ); \notag \\
   & \rm{Logistic}    & L_{\rm{log}}(\bx_i, y_i) &= \log( 1 + \exp( -
   \rho_{i,r} )).  \notag
\end{align*}
The two losses behave similarly for positive margin but differently for negative margin.
$L_{\rm{log}}$ has been reported to be more robust against outliers and misspecified data compared to $L_{\rm{exp}}$.
In the rest of this section, we present a coordinate descent based multi-class boosting as an approximate $\ell_1$-regularized fitting.
We then illustrate the similarity between both approaches.
Finally, we discuss various strategies that can be adopted to prevent over-fitting.

\subsection{Stage-wise multi-class boosting}

In this section, we design an efficient learning algorithm which maximizes the margin
of our training examples, $\rho_r(\bx_i, y_i), r = 1, \cdots, k$.
The general $\ell_1$-regularized optimization problem we want to solve is
\begin{align}
    \label{EQ:stage1}
        \min_{ \bW }   \quad
        {\textstyle \sum_{i=1}^m} {\textstyle \sum_{r=1}^k}
        L(\bx_i, y_i) + \nu  \| \bW \|_{1}, \quad   %
        \st  \bW \geq 0.           %
\end{align}
Here $L$ can be any convex loss functions and parameter $\nu$ controls the trade off between model complexity and small error penalty.
Although \eqref{EQ:stage1} is $\ell_1$-norm regularized, it is possible to design our algorithm with other $\ell_p$-norm regularized.
We first derive the Lagrange dual problems of the optimization with
both exponential loss and logistic loss, and propose our new stage-wise multi-class boosting.

\paragraph{Exponential loss}
The learning problem for an exponential loss can be written as,
\begin{align}
    \label{EQ:exp1}
        \min_{ \bW }   \quad
        &
        {\textstyle \sum_{i=1}^m} {\textstyle \sum_{r=1}^k}
        \exp \left[ - \left( \bHi \bwyi - \bHi \bwr  \right) \right] +
            \nu  \| \bW \|_{1},  \\ \notag
        \st \quad &
        \bW \geq 0;           %
\end{align}
We introduce auxiliary variables, $\rho_{i,r}$, and rewrite the primal problem as,
\begin{align}
    \label{EQ:exp2}
        \min_{ \bW, \brho}   \quad
        &
        \log \Bigl( \sum\nolimits_{i,r} \exp \left( -\rho_{i,r} \right)  \Bigr) +
            \nu  \| \bW \|_{1},   \\ \notag
        \st \quad &
        \rho_{i,r} =  \bHi \bwyi - \bHi \bwr, \forall i,
        \forall r, \bW \geq 0;           %
\end{align}
where $(i,r)$ represents the joint index through all of the data and all of the classes.
Here we work on the logarithmic version of the original cost function.
Since $\log(\cdot)$ is strictly monotonically increasing, this does not change
the original optimization problem. Note that the regularization
parameters in these two problems should have different values.
Here we introduce the auxiliary variable $\brho$ in order to arrive at the dual
problem that we need.
The Lagrangian of \eqref{EQ:exp2} can be written as,
\begin{align*}
    \Lag_{\exp} (\bW,&\brho,\bU,\bZ) \; = \log \Bigl( {\textstyle  \sum_{i,r}}
         \exp( - \rhoir ) \Bigr) +
        \nu {\textstyle  \sum_{r}} \bw_{:r} \\ \notag
      &- {\textstyle \sum_{i,r}} u_{ir}   (
      \rho_{i,r} - \bHi \bwyi
      + \bHi \bwr )
      - \trace(\bZ^\T \bW),
\end{align*}
with $\bZ \geq 0$.
To derive the dual, we have
\begin{align*}
    \bar{\Lag}_{\exp} (\bU,\bZ) \; &= \inf_{\bW,\brho} \Lag_{\exp} (\bW,\brho,\bU,\bZ) \\ \notag
    &= - \sup_{\rhoir} \Bigl( \uir \rhoir -
      \log( {\textstyle \sum_{i,r} } \exp(-\rhoir)) \Bigr) \\ \notag
      & + \inf_{\bW}
      \underbrace{
           \sum_{i,r} u_{ir} \bHi \bwyi
         - \sum_{i,r} u_{ir} \bHi \bwr + g(\bW) }_{\textrm{must be zero}}
\end{align*}
where $g(\bW) = \nu \sum_{r} \bw_{:r} - \trace(\bZ^\T \bW)$.
At optimum the first derivative of the Lagrangian with respect to
each row of $\bW$ must be zeros, \ie,
$\frac{\partial \Lag}{\partial \bw_{:r} } =  {\bf 0}$, and therefore
\begin{align}
    \notag
    \;
    \sum_{i \mid \yi=r}
         & \left( {\textstyle \sum_{l} } u_{i,l}  \right) \bHi
        -  \sum_{i} \uir \bHi
        = \bz_{r:}^\T - \nu \b1^\T  \\ \notag
    &\Rightarrow
    \sum_{i} \delta_{r,\yi} \left( \textstyle \sum_{l} u_{i,l} \right) \bHi
    - \sum_{i} \uir  \bHi
    \geq  -\nu \b1^\T, \forany r;
\end{align}
where
$\delta_{s,t}$ denotes the indication operator such that
$\delta_{s,t} = 1$ if $s = t$ and $\delta_{s,t} = 0$, otherwise.
Since the convex conjugate of the log-sum-exp function is
the negative entropy function. Namely,
the convex conjugate of $f(\ba) = \log \bigl(\sum_{i=1}^{m} \exp( a_i ) \bigr)$
is $f^{\ast}(\bb) = \sum_{i=1}^{m} b_i \log b_i$ if $\bb \geq {\bf 0}$ and
$\b1^\T  \bb = 1$; otherwise $f^{\ast} = \infty$.
The Lagrange dual problem can be derived as,
\begin{align}
    \label{EQ:exp3}
        \min_{ \bU }   \quad
        &
        {\textstyle \sum_{i,r}} u_{ir} \log( u_{ir} ), \\ \notag
        \st \quad &
        {\textstyle \sum_{i}} \left[ \delta_{r, \yi}
            \left( {\textstyle \sum_{l=1}^k} u_{il} \right) - u_{ir} \right] \bHi \leq \nu {\boldsymbol 1}^\T, \\ \notag
        \quad &
        {\textstyle \sum_{i,r}} u_{ir} = 1, \bU \geq 0.
\end{align}
Note that the objective function of the dual encourages the dual variables,
$\bU$, to be uniform.

\paragraph{Logistic loss}
The learning problem of logistic loss can be expressed as,
\begin{align}
    \label{EQ:log1}
        \min_{ \bW, \brho}  \quad
        &
        \sum\nolimits_{i,r}
            \log \bigl( 1 + \exp \left( -\rho_{i,r} \right) \bigr) +
            \nu  \| \bW \|_{1},   \\ \notag
            \st \quad &
            \rho_{i,r} =  \bHi \bwyi - \bHi \bwr, \forall i,
            \forall r, \bW \geq 0;           %
\end{align}
The Lagrangian of \eqref{EQ:log1} can be written as,
\begin{align*}
    \Lag_{\log} (\bW,&\brho,\bU,\bZ) \; =
    {\textstyle \sum_{i,r}} \log \bigl( 1 + \exp( - \rhoir) \bigr) +
        \nu {\textstyle  \sum_{r}} \bw_{:r} \\ \notag
      &- {\textstyle \sum_{i,r}} u_{ir}   (
      \rho_{i,r} - \bHi \bwyi
      + \bHi \bwr )
      - \trace(\bZ^\T \bW),
\end{align*}
with $\bZ \geq 0$.
Following the above derivation and
using the fact that the conjugate of logistic loss
$f(a) = \log \bigl( 1 + \exp(-a) \bigr)$ is
$f^{\ast}(b) = (-b) \log(-b) + (1+b) \log(1+b)$,
if $-1 \leq b \leq 0$; otherwise $f^{\ast}(b) = \infty$.
The Lagrange dual\footnote{Note that the sign of $\bU$ has been reversed.} can be written as,
\begin{align}
    \label{EQ:log2}
        \min_{ \bU }   \quad
        &
        {\textstyle \sum_{i,r}} \bigl[ u_{ir} \log( u_{ir} ) + (1 - u_{ir}) \log( 1 - u_{ir} ) \bigr], \\ \notag
        \st \quad &
        {\textstyle \sum_{i}} \left[ \delta_{r, \yi}
            \left( {\textstyle \sum_{l=1}^k} u_{il} \right) - u_{ir} \right] \bHi \leq \nu {\boldsymbol 1}^\T, \forall r; \\ \notag
        \quad &
        {\textstyle \sum_{i,r}} u_{ir} = 1, \bU \geq 0.
\end{align}
Since both \eqref{EQ:exp2} and \eqref{EQ:log1} are convex, both
problems are feasible and the Slater's conditions
are satisfied, the duality gap between the primal, \eqref{EQ:exp2} and \eqref{EQ:log1},
and the dual, \eqref{EQ:exp3} and \eqref{EQ:log2}, is zero.
Therefore, the solution of \eqref{EQ:exp2} and \eqref{EQ:exp3},
and \eqref{EQ:log1} and \eqref{EQ:log2} must be the same.
Although \eqref{EQ:exp3} and \eqref{EQ:log2} have identical constraints,
we will show later that their solutions
(selected weak classifiers and coefficients) are different.

\paragraph{Finding weak classifiers}
\revised{
From the dual, the set of constraints can be infinitely large, \ie,
\begin{align}
    \label{EQ:weak1}
    {\textstyle \sum_{i}} \left[ \delta_{r, \yi}
            \left( {\textstyle \sum_{l=1}^k} u_{il} \right) - u_{ir} \right] \hbar(\bxi) \leq \nu, \forany r, \forany \hbar(\cdot) \in \mathcal{H}.
\end{align}
For decision stumps, the size of $\mathcal{H}$ is
the number of features times the number of samples.
For decision tree, the size of $\mathcal{H}$ would grow exponentially
with the tree depth.
Similar to LPBoost, we apply a technique known as column generation to identify an optimal set of constraints\footnote{Note that constraints in the dual correspond to variables in the primal. An optimal set of constraints in the dual would correspond to a  set of variables in the primal that we are interested in.} \cite{Demiriz2002LPBoost}.
The high-level idea of column generation is to only consider a small subset
of the variables in the primal, \ie, only a subset of $\bW$ is
considered.
The problem solved using this subset is called
the restricted master problem (RMP).
At each iteration, one column, which corresponds to a variable in
the primal or a constraint in the dual, is added and
the restricted master problem is solved to obtain both primal
and dual variables.
We then identify any violated constraints which we have not added to the
dual problem.
These violated constraints correspond to variables in primal
that are not in RMP.
If no single constraint is violated, then we stop
since we have found the optimal dual solution to the original problem
and we have the optimal primal/dual pair.
In other words, solving the restricted problem is equivalent to
solving the original problem.
Otherwise, we append this column to the restricted master problem
and the entire procedure is iterated.
Note that any columns that violate the dual feasibility can be added.
However, in order to speed up the convergence, we add the most violated
constraint at each iteration.
In our case, the most violated constraint corresponds to:
\begin{align}
    \label{EQ:weak2}
    \hbar^{\ast}(\cdot) = \argmax_{\hbar(\cdot)\in \mathcal{H}, r}
    \;
    {\textstyle \sum}_{i=1}^m \left[ \delta_{r, \yi}
            \left( {\textstyle \sum_{l=1}^k} u_{il} \right) - u_{ir} \right] \hbar(\bxi).
\end{align}
Solving this subproblem is identical to finding a weak classifier with
minimal weighted error in AdaBoost (since dual variables, $\bU$, can be viewed as sample weights).
At each iteration, we add the most violated constraint into the dual problem.
The process continues until we can not find any violated constraints.

Through Karush-Kunh-Tucker (KKT) optimality condition, the gradient of
Lagrangian over primal variable, $\brho$, and dual variable, $\bU$, must vanish at the optimal.
Let $(\bW^{\ast}, \brho^{\ast})$ and $(\bU^{\ast}, \bZ^{\ast})$ be any primal
and dual optimal points with zero duality gap.
One of the KKT conditions tells us that
$\nabla_{\brho} \Lag_{\exp} (\bW^{\ast}, \brho^{\ast}, \bU^{\ast}, \bZ^{\ast}) = 0$
and $\nabla_{\brho} \Lag_{\log} (\bW^{\ast}, \brho^{\ast}, \bU^{\ast}, \bZ^{\ast}) = 0$.
We can obtain the relationship between the optimal primal and dual variables as,
\begin{align}
    \label{EQ:KKT1}
    & {\rm Exponential} \quad & u^{\ast}_{ir} &= \frac{ \exp(-\rho^{\ast}_{i,r}) }{ \sum_{i,r} \exp( -\rho^{\ast}_{i,r} )}, \\
    \label{EQ:KKT2}
        & {\rm Logistic} \quad & u^{\ast}_{ir} &= \frac{ \exp(-\rho^{\ast}_{i,r}) }{
             1 + \exp( -\rho^{\ast}_{i,r} )  }.
\end{align}
}

\begin{figure*}[t]
    \begin{center}
        \includegraphics[width=0.275\textwidth,clip]{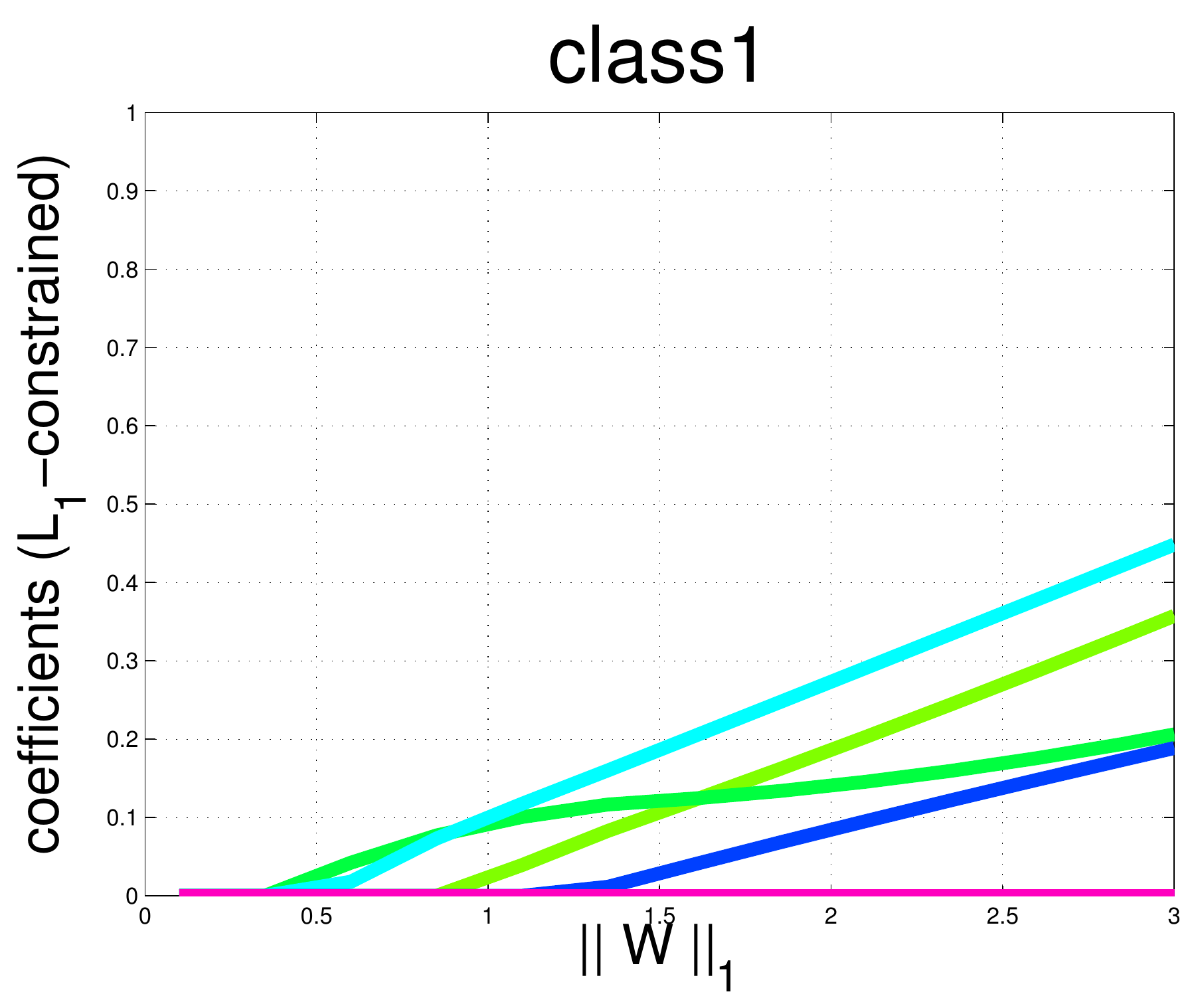}
        \includegraphics[width=0.25\textwidth,clip]{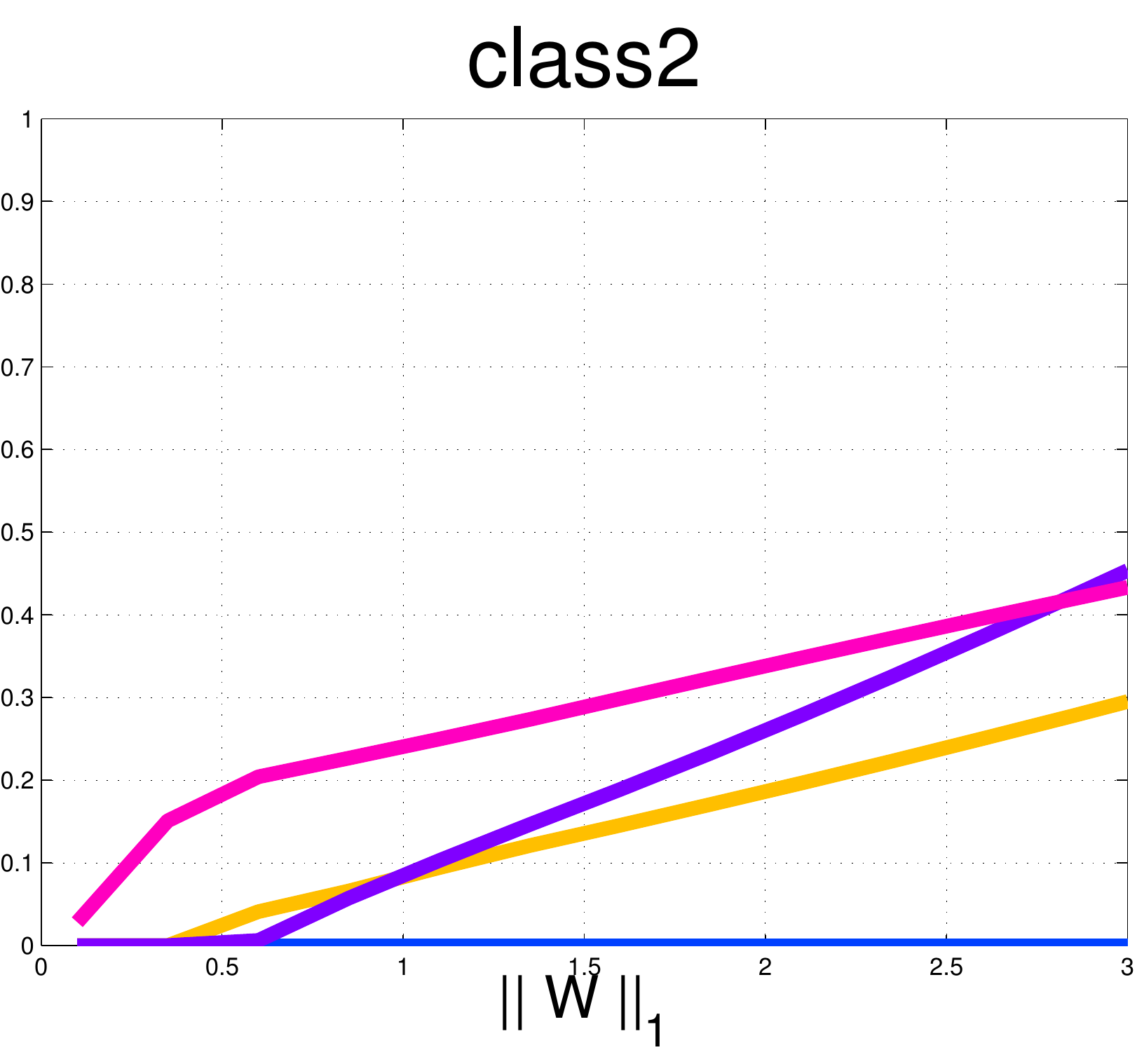}
        \includegraphics[width=0.25\textwidth,clip]{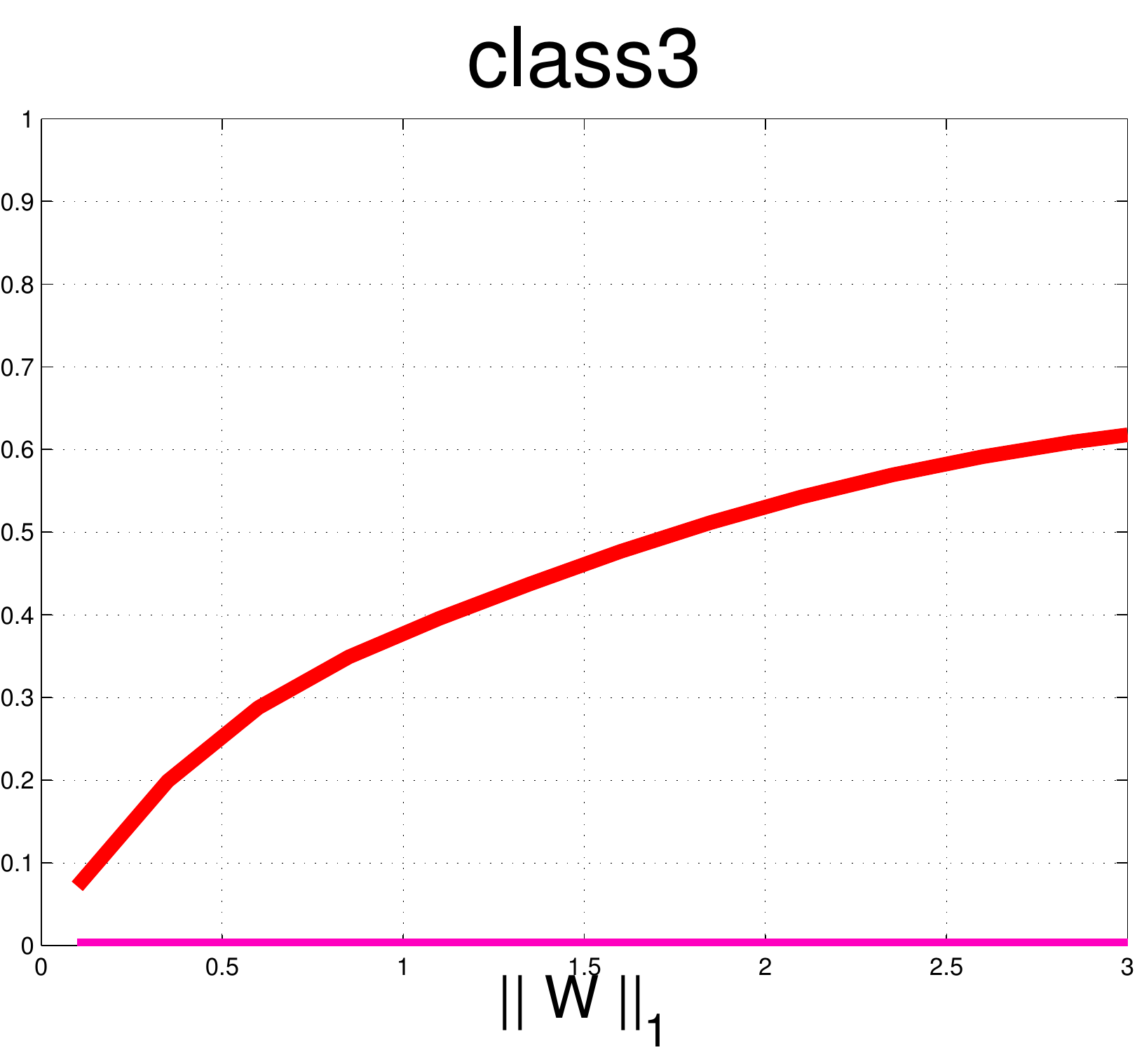}
        \includegraphics[width=0.27\textwidth,clip]{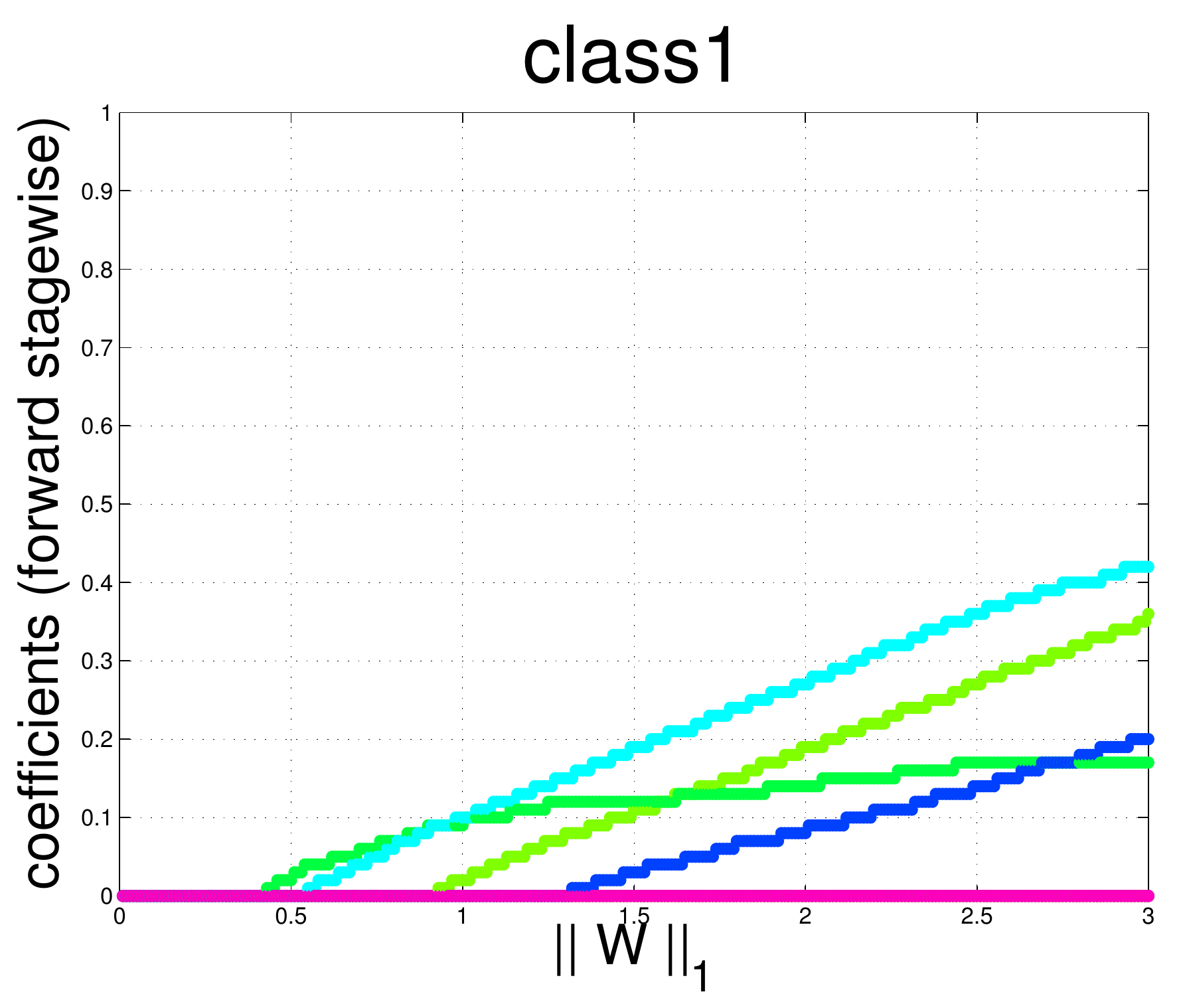}
        \includegraphics[width=0.25\textwidth,clip]{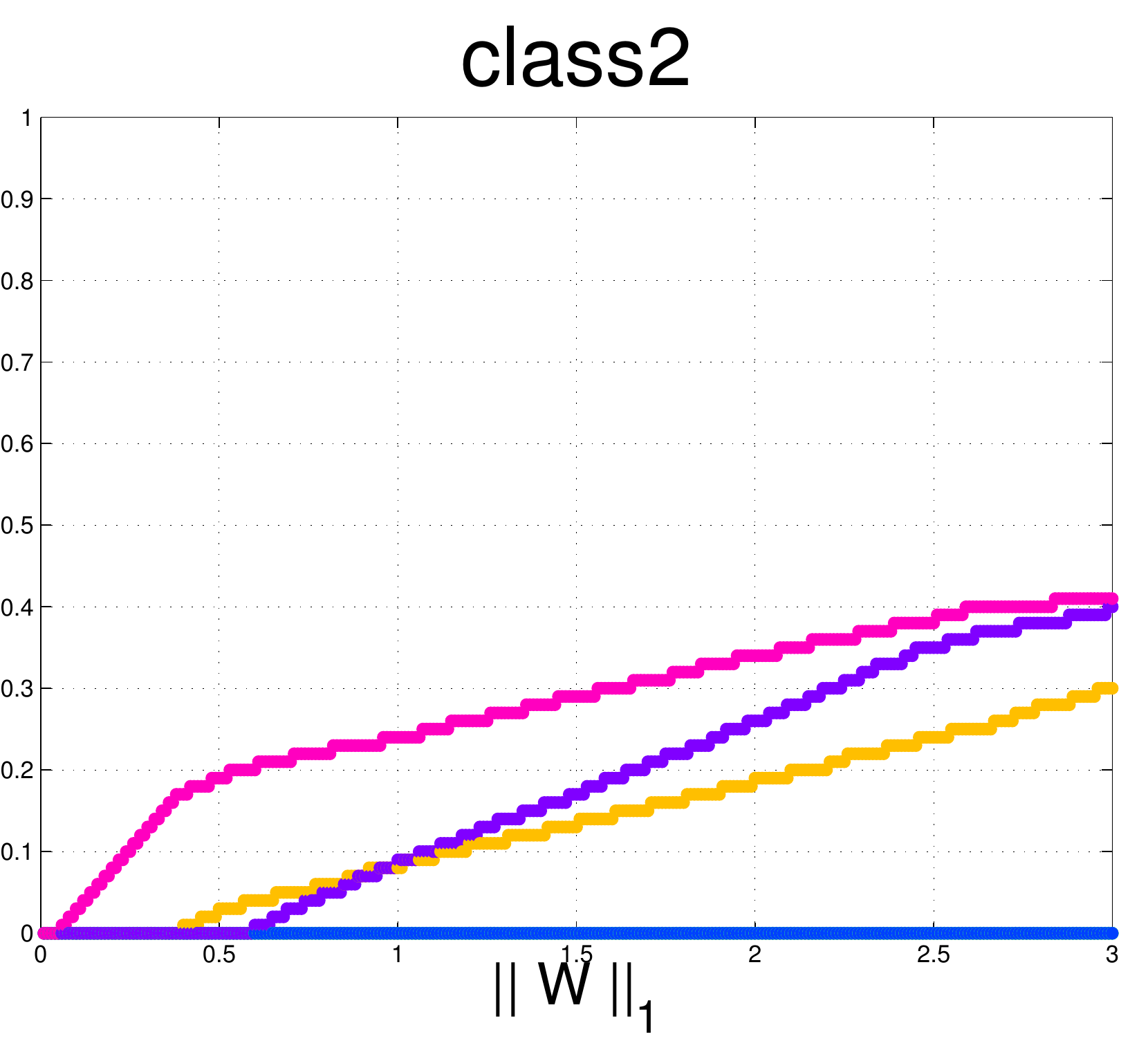}
        \includegraphics[width=0.25\textwidth,clip]{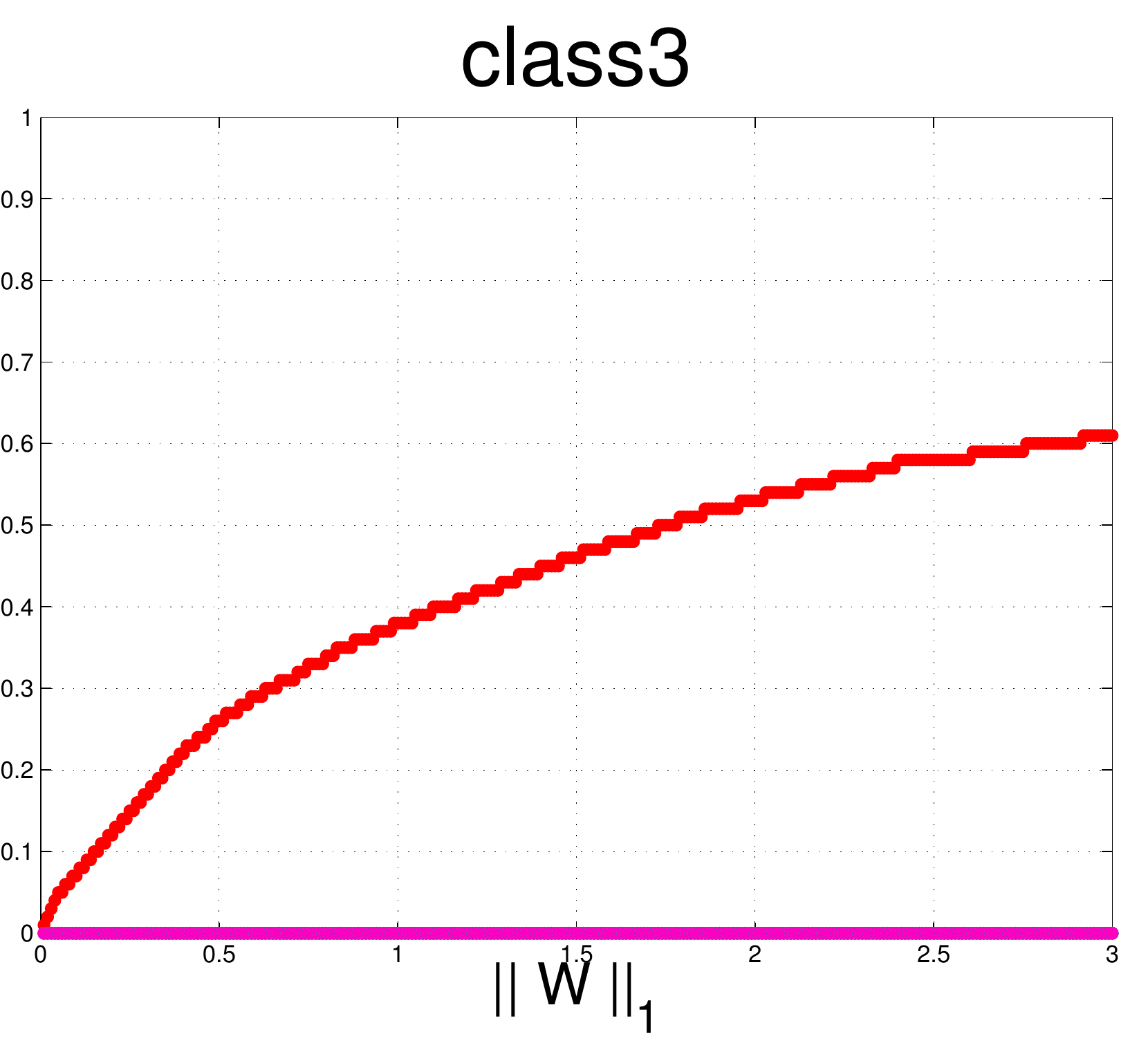}
        \includegraphics[width=0.275\textwidth,clip]{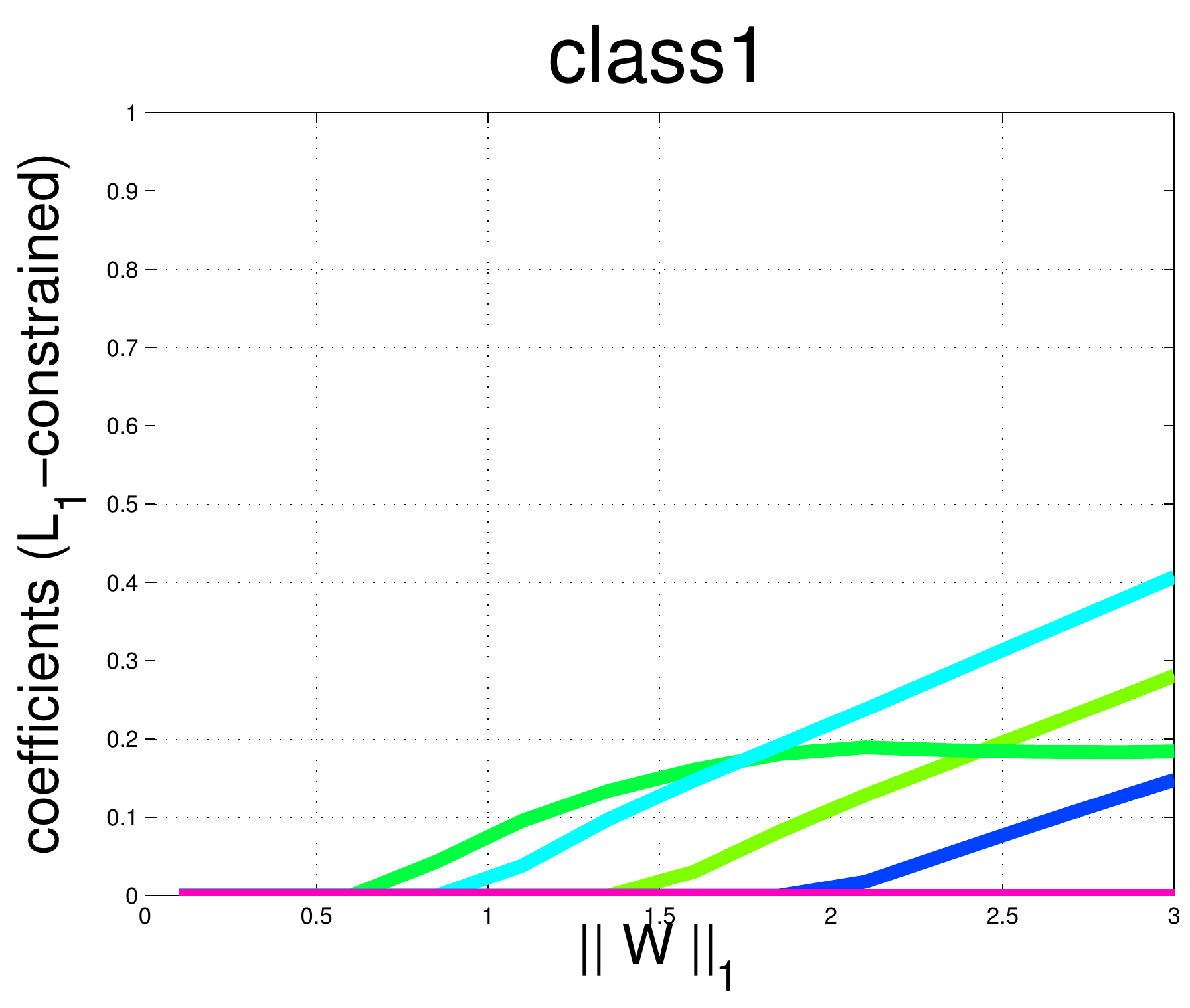}
        \includegraphics[width=0.25\textwidth,clip]{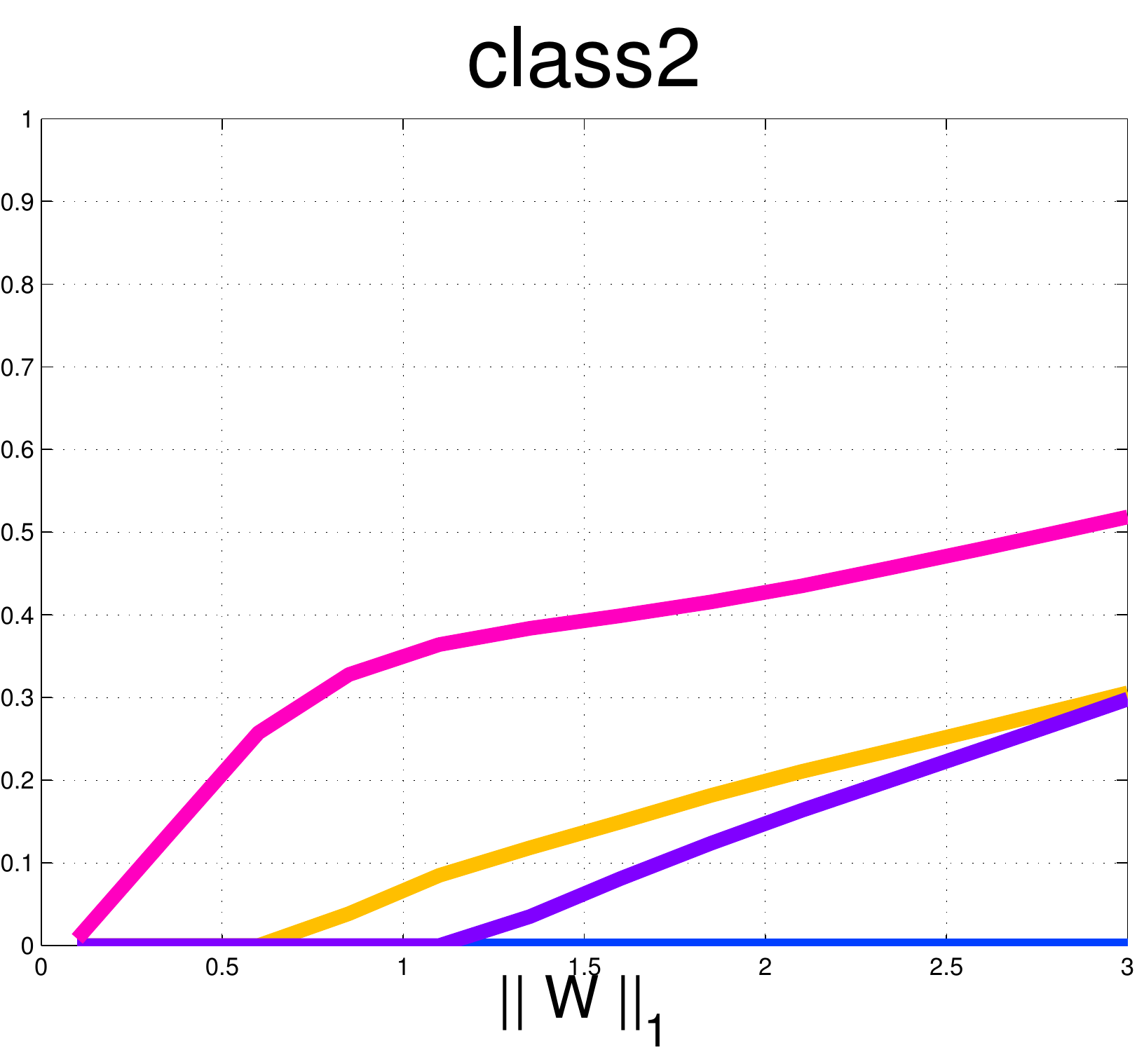}
        \includegraphics[width=0.25\textwidth,clip]{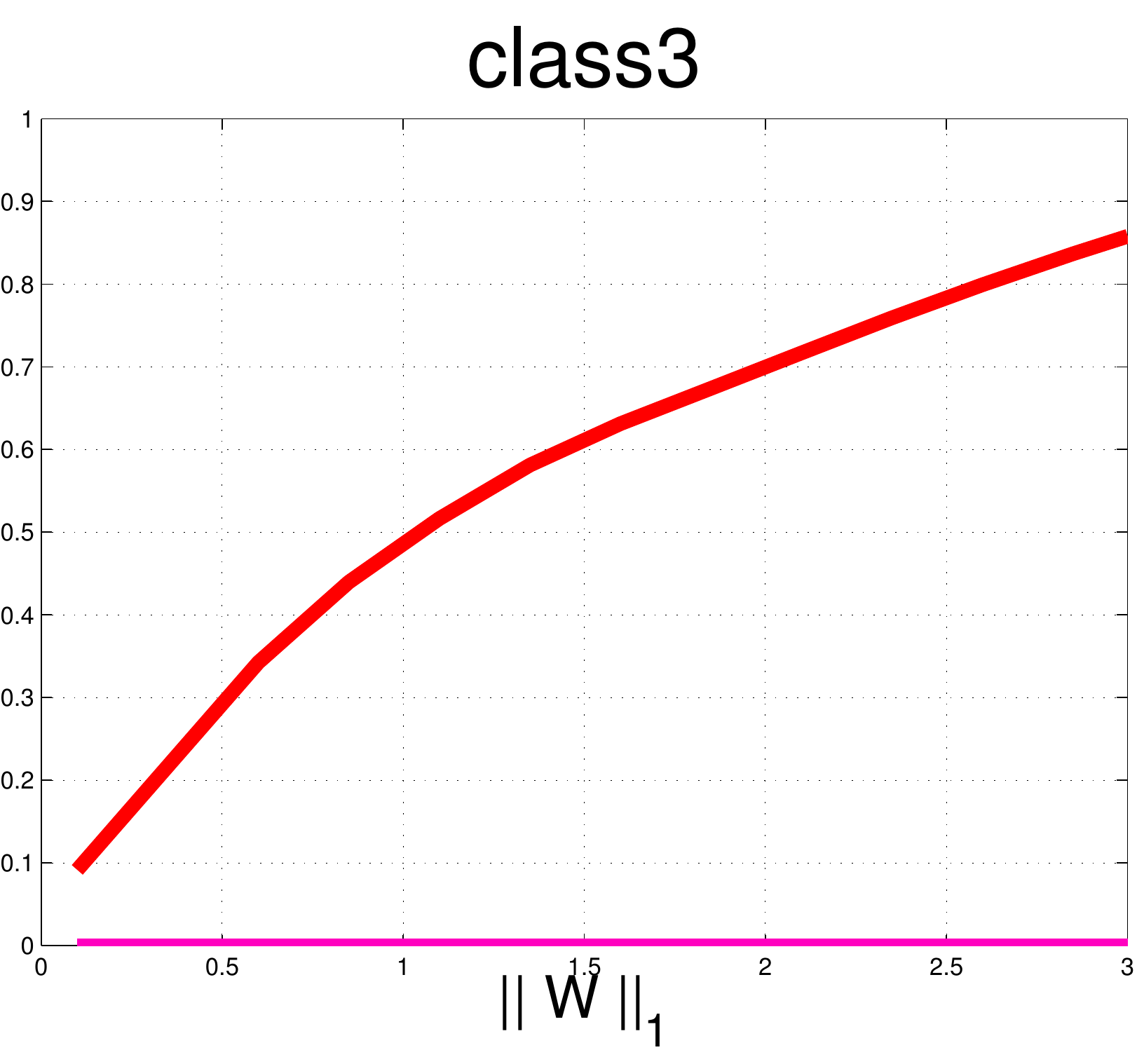}
        \includegraphics[width=0.27\textwidth,clip]{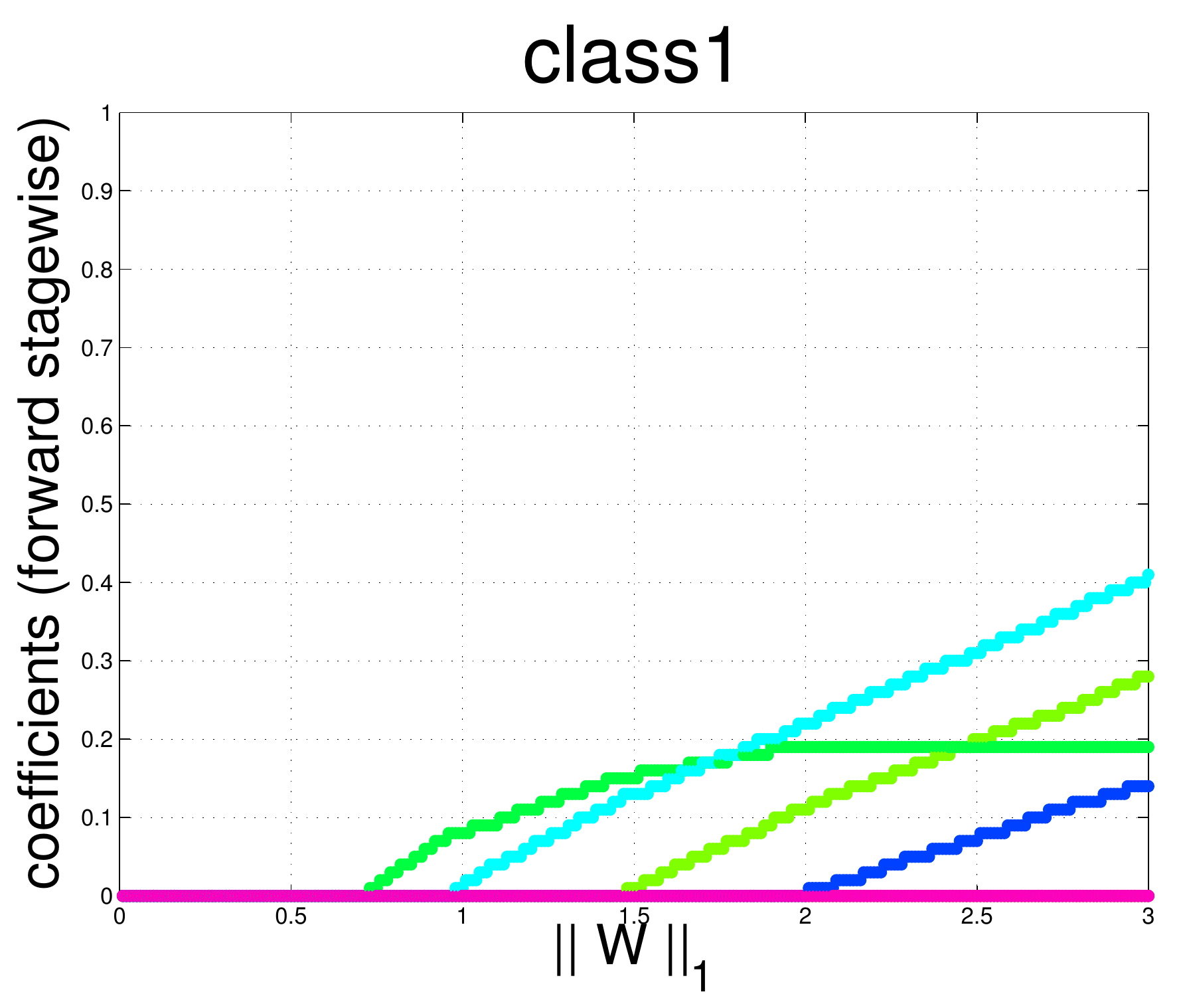}
        \includegraphics[width=0.25\textwidth,clip]{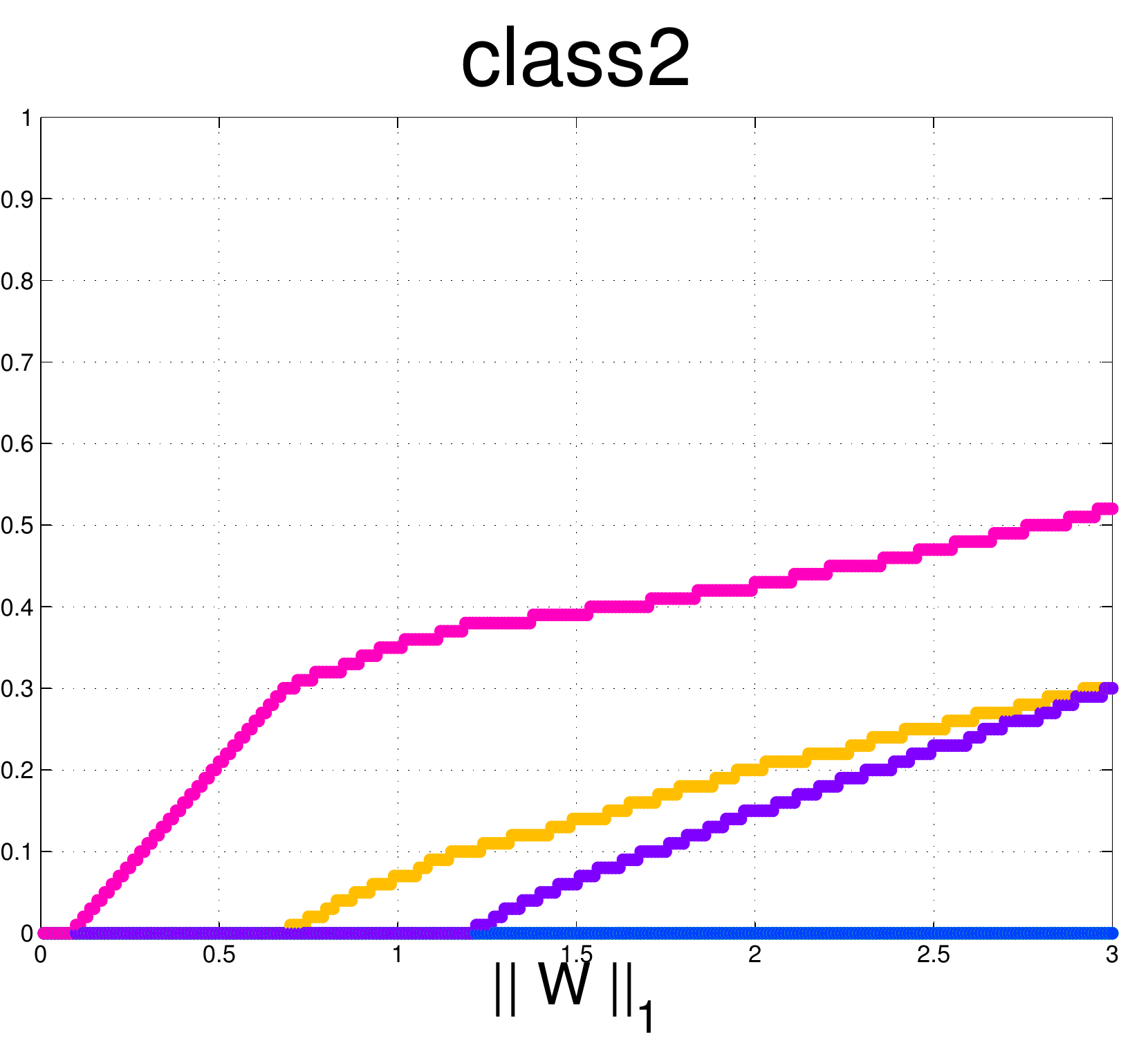}
        \includegraphics[width=0.25\textwidth,clip]{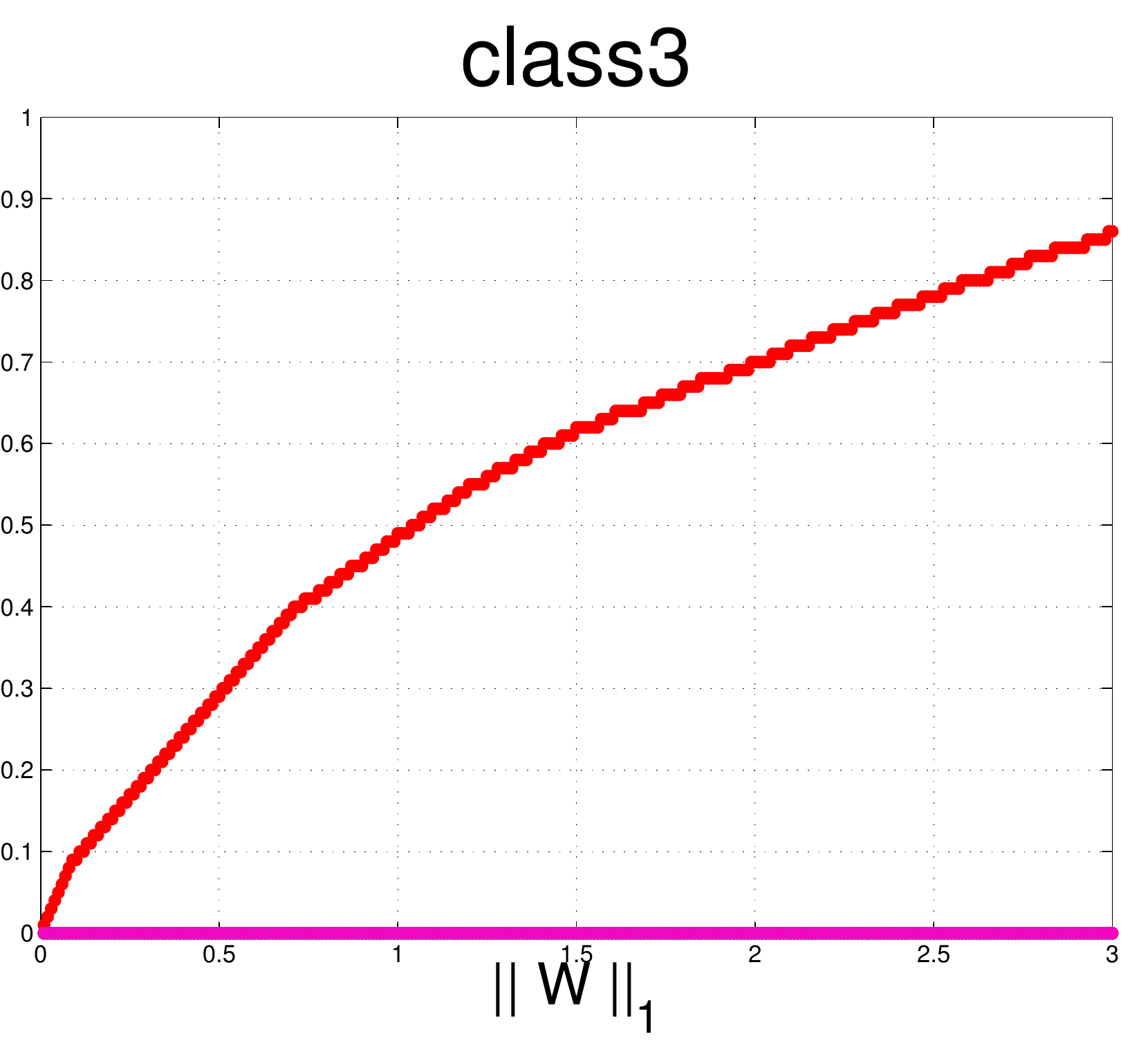}
    \end{center}
    \caption{
    Best viewed in color.
    {\em Top:} Exact coefficient paths for $\ell_1$-constrained {\bf exponential loss}
    {\em Second row:} Coefficient paths of our stage-wise boostings ({\bf exponential loss})
    {\em Third row:} Exact coefficient paths for $\ell_1$-constrained {\bf logistic loss} \eqref{EQ:l1}
    {\em Bottom:} Coefficient paths of our stage-wise boostings ({\bf logistic loss}).
    We train multi-class classifiers on USPS data sets (digit $3$,$6$ and $9$).
    Each figure corresponds to each digit and each curve in the figure corresponds to their coefficients value.
    Note the similarity between coefficients of both algorithms.
    }
    \label{fig:sw_l1}
\end{figure*}

\paragraph{Optimizing weak learners' coefficients}
Weak learners' coefficients can be calculated in a totally corrective
manner as in \cite{Shen2011Direct}.
However, the drawback of \cite{Shen2011Direct} is that
 the training time is often slow when the number of training samples and classes are large
 because the primal variable, $\bW$, needs to be updated at every boosting iteration.
In this paper, we propose a more efficient approach based on a stage-wise algorithm similar to those derived in AdaBoost.
The advantages of our approaches compared to \cite{Shen2011Direct} are
1) it is computationally efficient as we only update weak learners' coefficient at the current iteration and
2) our method is less sensitive to the choice of the regularization parameters and, as a result,
the training time can be much simplified since we no longer have to cross-validate these parameter.
We will show later in our experiments that
the regularization parameter only needs to be set to a {sufficiently
small} value
to ensure good classification accuracy.
By inspecting the primal problem, \eqref{EQ:exp2} and \eqref{EQ:log1}, the optimal $\bW$ can be calculated analytically as follows.
At iteration $t$, where $1 \leq t \leq n$, we fix the value of $\bw_{1:}$, $\bw_{2:}$,
$\cdots$, $\bw_{t-1,:}$.
So $\bw_{t:}$ is the only variable to be optimized.
The primal cost function for exponential loss can then be written as,
\begin{align}
    \label{EQ:coeff_sw1}
        \min_{ \bw_{t:} }   \quad
        &
        \log \Bigl( {\textstyle \sum_{i=1}^m}   {\textstyle  \sum_{r=1}^k}
        u_{ir}^{t-1} \exp \left( -  \rho_{i,r}^t   \right) \Bigr) +
            \nu  \| \bw_{t:} \|_{1}, \quad    \\ \notag
        \st \quad &
        \bw_{t:} \geq 0;           %
\end{align}
where $u_{ir}^{t-1} = \exp \left( -  {\textstyle \sum_{j=1}^{t-1}} \rho_{i,r}^j  \right) $ and
$\rho_{i,r}^j = \hbar_j(\bxi) w_{j\yi} - \hbar_j(\bxi) w_{jr}$.
Here we drop the terms that are irrelevant to $\bw_{t:}$ and
$u_{ir}^0$ is initialized to $\frac{1}{mk}$.
At each iteration, we compute $u_{ir}^t$ and cache the value for the next iteration.
Similarly, the cost function for logistic loss is,
\begin{align}
    \label{EQ:coeff_sw2}
        \min_{ \bw_{t:} }   \quad
        &
        {\textstyle  \sum_{i=1}^m }  {\textstyle   \sum_{r=1}^k }
        \log \Bigl( 1 + u_{ir}^{t-1} \exp \left( -  \rho_{i,r}^t   \right) \Bigr) +
            \nu  \| \bw_{t:} \|_{1},    \\ \notag
        \st \quad &
        \bw_{t:} \geq 0;           %
\end{align}
The above primal problems, \eqref{EQ:coeff_sw1} or \eqref{EQ:coeff_sw2}, can be solved using an efficient Quasi-Newton method like L-BFGS-B, and the
dual variables can be obtained using the KKT condition, \eqref{EQ:KKT1} or \eqref{EQ:KKT2}.
The details of our multi-class stage-wise boosting algorithm are given in Algorithm~\ref{ALG:alg1}.

\input{alg1.tex}

\paragraph{Computational complexity}
In order to appreciate the performance gain, we briefly analyze the complexity of the new approach and MultiBoost \cite{Shen2011Direct}.
The time consuming step in Algorithm~\ref{ALG:alg1} is in step \cone\ (weak classifier learning) and \cfour\ (calculating coefficients).
In step \cone, we train a weak learner by solving the subproblem \eqref{EQ:weak2}.
For simplicity, we use decision stumps as weak learners.
The fastest way to train the decision stump is to sort feature values and
scan through all possible threshold values sequentially to update \eqref{EQ:weak2}.
The algorithm takes $\bigO(m \log m)$ for sorting and $\bigO(km)$ for scanning $k$ classes.
At each iteration, we need to train $d$ decision stumps (since $\bx \in \Real^d$).
Hence, this step takes $\bigO(dm \log m + dkm)$ at each iteration.
In step \cfour, we solve $k$ variables at each iteration.
Let us assume the computational complexity of L-BFGS is roughly $\bigO(n^{2.5})$.
The algorithm spends $\bigO(k^{2.5})$ at each iteration.
Hence, the total time complexity for $t$ boosting iterations is $\bigO(tdm \log m + tdkm + tk^{2.5})$.
Roughly, the first term dominates when the number of samples is large and
the last term dominates when the number of classes is large.

We also analyze the computational complexity during training of MultiBoost.
The time complexity to learn weak classifiers in their approach would be the same as ours.
However, in step \cfour, they would need to solve $k \cdot t$
variables (since the algorithm is fully corrective).
The time complexity for this step\footnote{We have $1^2 + 2^2 + \cdots + n^2 = O(n^3)$ and
$1^3 + 2^3 + \cdots + n^3 = O(n^4)$.}
is $\bigO(k^{2.5} + (2k)^{2.5} + \cdots + (tk)^{2.5})$ $>$
$\bigO(k^{2.5} t^3)$.
Hence, the total time complexity for MultiBoost is $> O(tdm \log m + tdkm + k^{2.5}t^3)$.
Clearly, the last term will dominate when the number of iterations is large.
For example, training a multi-class classifier with $100$ samples,
$100$ features, $10$ classes for $1000$ iterations using our approach
would require $\bigO(10^8)$ while MultiBoost would require
$\bigO(10^{11})$.
{\em For this simple scenario,
our approach already speeds up the training time by three orders of magnitudes.}

\subsection{Discussion}

\paragraph{Binary classification}
Here we briefly point out the connection between our multi-class formulation and binary classification algorithms such as AdaBoost.
We note that AdaBoost sets the regularization parameter, $\nu$ in \eqref{EQ:exp1}, to be
zero \cite{Shen2010Dual} and it minimizes the exponential loss function.
{\em
The stage-wise optimization strategy of AdaBoost implicitly enforces
the $ \ell_1 $ regularization on the coefficients of weak learners.
} See details in \cite{Shen2010Dual,Rosset2004Boosting}.
We can simplify our exponential loss learning problem, \eqref{EQ:exp1}, for a binary case ($k = 2$) as,
\begin{align}
    \notag
        & \min_{ \bW }   \;
        {\textstyle \sum_{i=1}^m} {\textstyle \sum_{r=1}^2}
        \exp \left[ - \left( \bHi \bwyi - \bHi \bwr  \right) \right]
        \; \st  \bW \geq 0;
        \\
        \label{EQ:binary1}
        =& \min_{ \balpha }  \;
        {\textstyle \sum_{i=1}^m}
        \exp \left( - z_i  \bHi \balpha  \right)
        , \quad \st  \balpha \geq 0,
\end{align}
where $z_i = 1$ if $y_i = 1$, $z_i = -1$ if $y_i = 2$ and $\balpha = \bw_{:1} - \bw_{:2}$.
AdaBoost minimizes the exponential loss function via coordinate descent.
At iteration $t$, Adaboost fixes the value of $\alpha_1, \alpha_2, \cdots, \alpha_{t-1}$ and solve $\alpha_t$.
So \eqref{EQ:binary1} can simply be simplified to,
\begin{align}
    \label{EQ:binary2}
        \min_{ \alpha_t }   \quad
        &
         {\textstyle \sum_{i=1}^m}
        u_{i}^{t-1} \exp \left( -  \rho_{i}^t   \right), \quad %
        \st %
        \alpha_t \geq 0;           %
\end{align}
where $u_{i}^{t-1} = \exp \left( {- \textstyle \sum_{j=1}^{t-1}} \rho_i^j \right)$ and $\rho_i^j = z_i h_j(\bxi) \alpha_j$.
By setting the first derivative of \eqref{EQ:binary2} to zero, a
closed-form solution of $\alpha_t$ is:
$\alpha_t^{\ast} = \frac{1}{2} \log \frac {1 - \epsilon_t  } { \epsilon_t }$
where $\epsilon_t = \sum_{i: z_i \neq \hbar_t(\bxi)} u_i^{t-1}$.
$\epsilon_t$ corresponds to a weighted error rate with respect to the distribution of dual variables.
By replacing step \cfour\ in Algorithm~\ref{ALG:alg1} with \eqref{EQ:binary1}, our approach would yield an identical solution to AdaBoost.

\paragraph{The $\ell_1$-constrained classifier and our boosting}
Rosset \etal pointed out that by setting the coefficient value to be
small, gradient-based boosting tends to follow the solution of
$\ell_1$-constrained maximum margin classifier, \eqref{EQ:l1}, as a
function of $\gamma$ under some mild conditions
\cite{Rosset2004Boosting}:
\begin{align}
    \label{EQ:l1}
    \bW^{\ast}(\gamma) =  \min_{ \bW }   \quad
        &
          {\textstyle  \sum_i } L(y_i, F(\bxi)), \quad \\ \notag
        \st \quad &
        \| \bW \|_{1} \leq \gamma, \bW \geq 0;
\end{align}
We conducted a similar experiment on our multi-class boosting to illustrate the similarity between our forward stage-wise boosting and the optimal solution of \eqref{EQ:l1} on USPS data set.
The data set consists of $256$ pixels.
We randomly select $20$ samples from $3$ classes ($3$, $6$ and $9$).
For ease of visualization and interpretation, we limit the number of available decision stumps to $8$.
We first solve \eqref{EQ:l1} using \CVX package\footnote{Note that to solve \eqref{EQ:l1}, the algorithm must access all weak classifiers a priori.} \cite{cvx}.
We then train our stage-wise boosting as discussed previously.
However, instead of solving \eqref{EQ:coeff_sw1} or \eqref{EQ:coeff_sw2}, we set the weak learner's coefficient of the selected class in \eqref{EQ:weak2} to be $0.01$ and the weak learner's coefficient of other classes to be $0$.
The learning algorithm is run for $1000$ boosting iterations.
The coefficient paths of each class are plotted in the second row in Fig.~\ref{fig:sw_l1} (the first three columns correspond to exponential loss and the the last three correspond to logistic loss).
We compare the coefficient paths for our boosting and $\ell_1$-constrained exponential loss and logistic loss in Fig.~\ref{fig:sw_l1}.
We observe that both algorithms give very similar coefficients.
This experimental evidence leads us to the connection between the solution of our multi-class boosting and the solution of \eqref{EQ:l1}.
Rosset \etal have also pointed this out for a binary classification problem \cite{Rosset2004Boosting}.
The authors incrementally increase the coefficient of the selected weak classifiers by a very small value and demonstrate that the final coefficient paths follow the $\ell_1$-regularized path.
In this section, we have demonstrated that our multi-class boosting also asymptotically converges to the optimal solution of $\ell_1$-regularized solution \eqref{EQ:l1}.

\paragraph{Shrinkage and bounded step-size}
In order to minimize over-fitting, strategies such as shrinkage \cite{Friedman2000Additive} and bounded step-size \cite{Zhang2005Boosting} can also be adopted here.
We briefly discuss each method and how they can be applied to our approach.
As discussed in previous section, at iteration $t$, we solve,
\begin{align}
    \bw_{t:}^{\ast} = \argmin_{\bw_{t:}} {\textstyle \sum_{i,r}}
        L \bigl( \yi, F^t (\bxi) +  \hbar_{t} ( \bxi ) w_{tr} \bigr). \notag
\end{align}
The alternative approach, as suggested by \cite{Friedman2000Additive},
is to shrink all coefficients to small values.
Shrinkage is simply another form of regularization. %
The algorithm replaces $\bw_{t:}$ with $\eta \bw_{t:}$ where $0 < \eta < 1$.
Since $\eta$ decreases the step-size, $\eta$ can be viewed as a learning rate parameter.
The smaller the value of $\eta$, the higher the overall accuracy as long as there are enough iterations.
Having a large enough iteration means that we can keep selecting the same weak classifier repeatedly if it remains optimal.
It is  observed in \cite{Friedman2000Additive} that shrinkage often produces a better generalization performance compared to line search algorithms.
Similar to shrinkage, bounded step-size can also be applied.
It caps $\bw_{t:}$ by a small value, \ie,
$w_{tr} = \min ( w_{tr}, \kappa )$ where $\kappa$ is often small.
The method decreases the step-size in order to provide a better generalization performance.

\begin{table*}
\caption{Average test errors with different values of $\nu$ in \eqref{EQ:stage1}.
  All experiments are repeated $50$ times with $500$ boosting iterations.
  The average error mean and standard deviation (shown in $\%$) are reported.
  The best average performance is shown in boldface.
  The first column displays the data set name and the number of classes.
  From the table, there is no significant difference between the final classification performance of various $\nu$ (as long as it is sufficiently small)
  }
  \centering
  \scalebox{1}
  {
  \begin{tabular}{l|ccccc|ccccc}
  \hline
    data &\multicolumn{5}{c|}{Exponential} & \multicolumn{5}{c}{Logistic}\\
  \cline{2-11}
    ($\#$ classes) & $\nu=0$ & $10^{-9}$ & $10^{-4}$ & $10^{-2}$ & $10^{-1}$ & $\nu=0$ & $10^{-9}$ &  $10^{-4}$ & $10^{-2}$ & $10^{-1}$ \\
  \hline
  \hline
  iris ($3$) &  $6.3$ ($3.4$) & $6.3$ ($3.4$) & $6.1$ ($3.5$) & $5.6$ ($2.8$) & $5.4$ ($3.2$) &
                $6.1$ ($3.5$) & $6.2$ ($3.7$) & $6.0$ ($3.2$) & $\mathbf{5.2}$ ($\mathbf{2.8}$) & $5.7$ ($3.5$) \\
  glass ($6$) &  $28.4$ ($7.0$) & $28.4$ ($7.0$) & $28.2$ ($6.4$) & $27.9$ ($6.1$) & $\mathbf{26.0}$ ($\mathbf{6.4}$) &
                $28.0$ ($7.6$) & $28.0$ ($7.6$) & $28.4$ ($7.1$) & $27.1$ ($7.2$) & $34.6$ ($6.2$) \\
  usps ($10$) & $\mathbf{9.1}$ ($\mathbf{2.4}$) & $\mathbf{9.1}$ ($\mathbf{2.4}$) & $9.8$ ($2.2$) & $13.0$ ($2.7$) & $29.0$ ($4.0$) &
                $9.1$ ($3.2$) & $9.1$ ($3.2$) & $9.6$ ($3.0$) & $12.4$ ($3.3$) & $27.5$ ($3.6$) \\
  pen ($10$) &  $\mathbf{6.4}$ ($\mathbf{2.4}$) & $\mathbf{6.4}$ ($\mathbf{2.4}$) & $6.8$ ($2.5$) & $9.8$ ($3.0$) & $36.8$ ($2.9$) &
                $6.4$ ($2.6$) & $6.5$ ($2.6$) & $6.5$ ($2.5$) & $8.0$ ($2.8$) & $26.9$ ($4.4$) \\
  news ($20$) & $53.6$ ($2.8$) & $53.6$ ($2.8$) & $53.4$ ($3.0$) & $56.3$ ($3.3$) & $95.9$ ($2.3$) &
                $\mathbf{53.3}$ ($\mathbf{3.0}$) & $\mathbf{53.3}$ ($\mathbf{3.0}$) & $53.5$ ($3.0$) & $56.4$ ($3.2$) & $77.6$ ($2.2$) \\
  \hline
  \end{tabular}
  }
  \label{tab:exp_reg}
\end{table*}

\begin{table}[t!]
  \caption{Average test errors with different shrinkage parameters, $\eta$.
  The average error mean and standard deviation (shown in $\%$) are reported.
  It can be seen in general, the test accuracy is improved in all data sets when shrinkage is applied.
  }
  \centering
  \scalebox{1}
  {
  \begin{tabular}{l|cccc}
  \hline
    data &\multicolumn{4}{c}{Exponential loss} \\
  \cline{2-5}
    ($\#$ classes)  & $\eta=1.0$ & $0.5$ & $0.2$ & $0.1$  \\
  \hline
  \hline
  iris ($3$)  &  $6.5$ ($3.5$) & $6.3$ ($3.4$) & $\mathbf{5.9}$ ($\mathbf{3.3}$) & $6.0$ ($3.2$) \\
  glass ($6$)  & $30.5$ ($6.6$) & $28.4$ ($7.0$) & $27.4$ ($7.1$) & $27.3$ ($6.4$)   \\
  usps ($10$)  &  $12.9$ ($2.8$) & $\mathbf{9.1}$ ($\mathbf{2.4}$) & $9.7$ ($2.4$) & $11.0$ ($2.7$) \\
  pen ($10$) &  $7.4$ ($2.8$) & $\mathbf{6.4}$ ($\mathbf{2.4}$) & $7.3$ ($2.8$) & $9.0$ ($3.1$) \\
  news ($20$) & $58.9$ ($2.9$) & $53.6$ ($2.8$) & $54.0$ ($3.2$) & $55.2$ ($3.0$)  \\
  \hline
  data & \multicolumn{4}{c}{Logistic loss}\\
  \cline{2-5}
    ($\#$ classes)  & $\eta=1.0$ & $0.5$ & $0.2$ & $0.1$ \\
  \hline
  \hline
  iris ($3$)  &  $6.4$ ($3.4$) & $6.2$ ($3.7$) & $\mathbf{5.9}$ ($\mathbf{3.3}$) & $\mathbf{5.9}$ ($\mathbf{3.3}$) \\
  glass ($6$)  & $29.2$ ($6.7$) & $28.0$ ($7.6$) & $26.7$ ($6.7$) & $\mathbf{26.3}$ ($\mathbf{6.4}$) \\
  usps ($10$)  & $10.8$ ($2.5$) & $9.1$ ($3.2$) & $9.5$ ($2.6$) & $10.9$ ($2.6$) \\
  pen ($10$) & $6.5$ ($2.5$) & $6.5$ ($2.6$) & $7.0$ ($2.8$) & $8.4$ ($3.1$) \\
  news ($20$) & $59.4$ ($3.8$) & $53.3$ ($3.0$) & $\mathbf{53.0}$ ($\mathbf{3.0}$) & $54.5$ ($3.2$) \\
  \hline
  \end{tabular}
  }
  \label{tab:exp_shrinkage}
\end{table}

\section{Experiments}
\label{sec:exp}

\subsection{Regularization parameters and shrinkage}
\revised{
In this experiment, we evaluate the performance of our algorithms on different
shrinkage parameters, $\eta$ and regularization parameters,
$\nu$ in \eqref{EQ:stage1}.
We investigate how shrinkage helps improve the generalization performance.
We use $5$ benchmark multi-class data sets.
We choose $50$ random samples from each class and
randomly split the data into two groups: $75\%$ for training and the rest for evaluation.
We set the maximum number of boosting iterations to $500$.
All experiments are repeated $50$ times.
We vary the value of $\eta$ between $0.1$ and $1.0$.
Experimental results are reported in Table~\ref{tab:exp_shrinkage}.
From the table, we observe a slight increase in generalization performances in all
data sets when shrinkage is applied.

In the next experiment, we evaluate how $\nu$ effects the final classification accuracy.
We experiment with $\nu$ in
$\{$$0$, $10^{-9}$, $10^{-4}$, $10^{-2}$, $10^{-1}$$\}$ using
both exponential loss and logistic loss.
Note that fixing $\nu$ is equivalent to selecting the maximum number of weak learners.
The iteration in our boosting algorithm continues until the algorithm can no longer
find the most violated constraint,
\ie, optimal solution has been found, or the maximum number of iterations is reached.
}
Table~\ref{tab:exp_reg} reports final classification errors.
From the table, we observe a similar classification accuracy when
$\nu$ is set to a small value ($0 \leq \nu \leq 10^{-4}$).
In this experiment, we do not observe over-fitting even when we set $\nu$ to $0$.
This is because the number of iterations serve as the regularization in our
problem.
For large $\nu$ ($\nu > 10^{-4}$), we observe that classification errors increase
as the number of classes increases.
Our conjecture is that as the classification problem becomes harder
(\ie, more number of classes),
the optimal $\hbar(\cdot)$ obtained in \eqref{EQ:weak2} would fail to
satisfy the stopping criterion for large $\nu$ (Step \ctwo\ in Algorithm~\ref{ALG:alg1}).
As a result, the algorithm terminates prematurely and poor performance is observed.
These experimental results demonstrate that
choosing a specific combination of $\nu$ and $\eta$ might not
have a strong influence on the final performance
($0.1 < \eta < 1.0$ and $\nu$ is sufficiently small).
However, one can cross-validate these parameters to achieve optimal results.
In the rest of our experiment, we apply a shrinkage value of $0.5$
and set $\nu$ to be $10^{-9}$.

\begin{table}[t!]
  \caption{Average test errors (in $\%$) between our approach and MultiBoost by varying the number of classes, training samples per class and boosting iterations.
  All algorithms perform similarly.
  }
  \centering
  \scalebox{1}
  {
  \begin{tabular}{l|cccc}
  \hline
  $\#$ classes &  $4$ & $8$ & $16$ & $26$ \\
  \hline
   \MultiEXPshort & $\mathbf{6.9}$ ($\mathbf{3.5}$) & $20.4$ ($3.4$)& $20.3$ ($2.8$) & $25.1$ ($2.0$) \\
   \MultiLOGshort & $\mathbf{6.9}$ ($\mathbf{3.5}$) & $\mathbf{20.0}$ ($\mathbf{3.2}$) & $\mathbf{19.9}$ ($\mathbf{3.0}$) & $\mathbf{24.9}$ ($\mathbf{2.0}$) \\
   MultiBoost \cite{Shen2011Direct} & $7.2$ ($3.2$) & $21.7$ ($3.3$) & $20.4$ ($3.1$) & $25.5$ ($2.5$) \\
  \hline
  $\#$ samples &  $20$ & $50$ & $100$ & $200$ \\
  \hline
  \MultiEXPshort &  $\mathbf{26.9}$ ($\mathbf{2.9}$) & $24.4$ ($1.9$)& $20.7$ ($1.3$)& $19.3$ ($1.0$) \\
  \MultiLOGshort &  $29.3$ ($3.2$) & $24.2$ ($2.1$) & $\mathbf{20.3}$ ($\mathbf{1.2}$)& $\mathbf{18.5}$ ($\mathbf{1.0}$) \\
   MultiBoost \cite{Shen2011Direct} & $29.9$ ($3.4$) & $\mathbf{24.1}$ ($\mathbf{2.5}$)& $20.5$ ($1.8$) & $19.8$ ($4.5$) \\
  \hline
  \end{tabular}
  }
  \label{tab:exp_multi1}
\end{table}

\begin{figure*}[t!]
    \begin{center}
        \includegraphics[width=0.30\textwidth,clip]{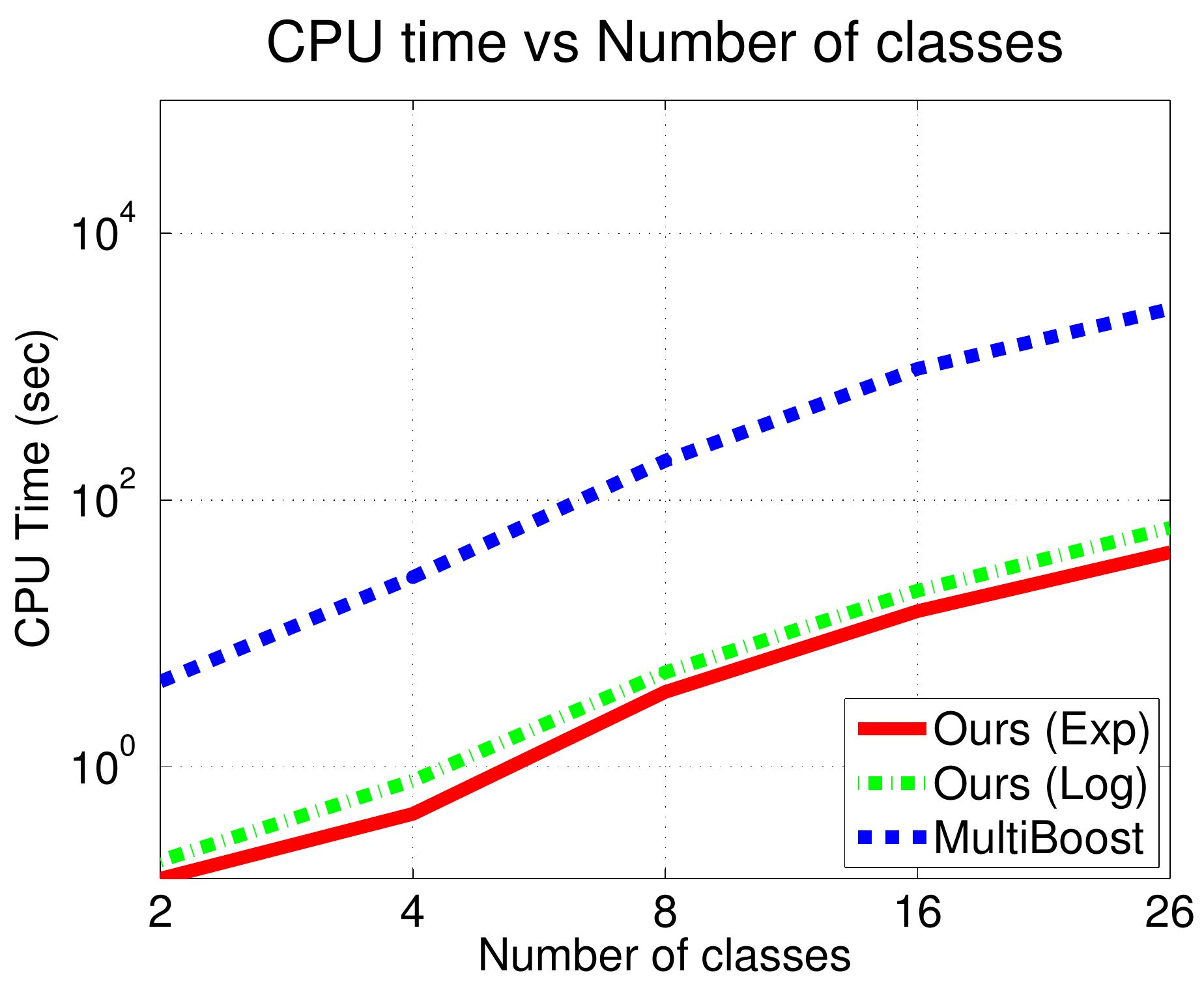}
        \includegraphics[width=0.30\textwidth,clip]{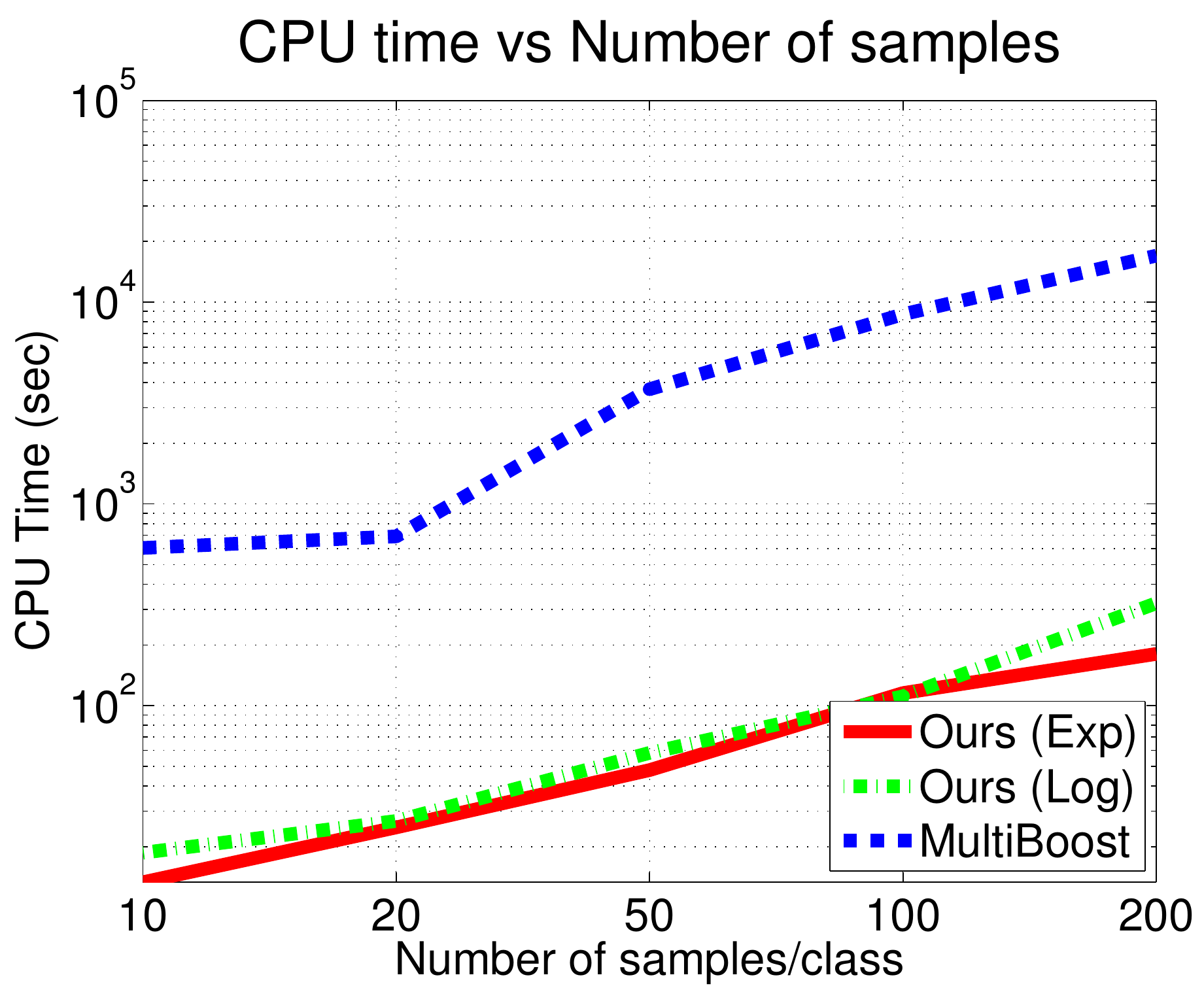}
        \includegraphics[width=0.30\textwidth,clip]{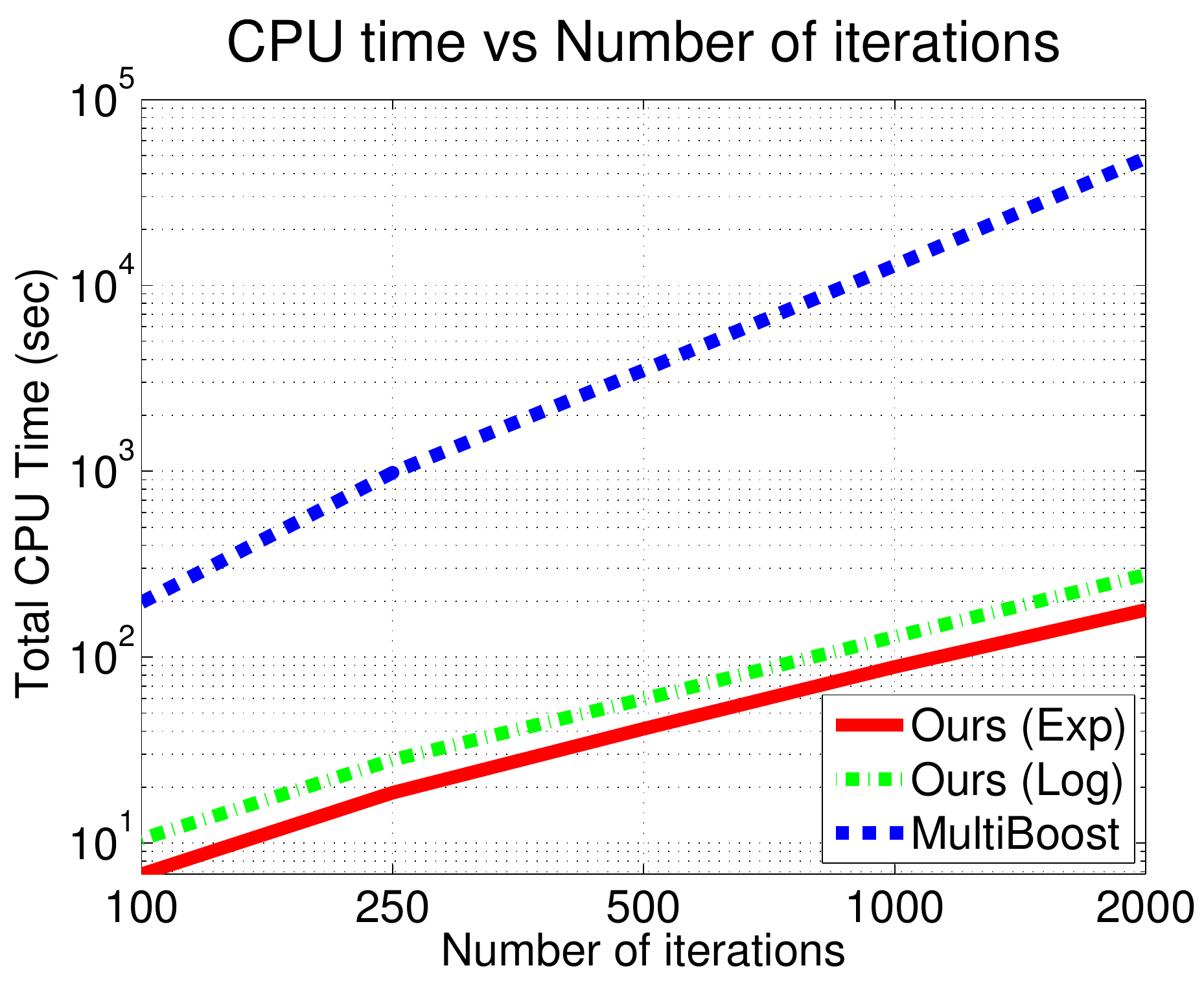}
    \end{center}
    \vspace{-.2cm}
    \caption{
    Average CPU time on a $\log$-$\log$ scale.
    The vertical axis corresponds to the required CPU time to calculate weak learners' coefficients (step \cfour\ in Algorithm~\ref{ALG:alg1}).
    In this figure, we vary the number of classes and boosting iterations.
    Our algorithm is at least two orders of magnitude faster than MultiBoost.
    }
    \label{fig:exp_multi}
\end{figure*}

\begin{table*}[t]
  \caption{Average test errors and their standard deviations (shown in $\%$)
  and the time to calculate $\bW$ for different algorithms.
  The best average performance is shown in boldface.
  Clearly, all algorithms perform comparably.
  However, the training time of our algorithms is much less than MultiBoost \cite{Shen2011Direct}.
  }
  \centering
  \scalebox{1}
  {
  \begin{tabular}{l|cc|cc|cc|cc}
  \hline
    data &\multicolumn{2}{c|}{\MultiEXPshort} & \multicolumn{2}{c|}{\MultiLOGshort} &\multicolumn{2}{c|}{MultiBoost \cite{Shen2011Direct}} & \multicolumn{2}{c}{Speedup factor} \\
  \cline{2-9}

  ($\#$ classes/$\#$ dims)  & Error  & Time & Error & Time &  Error & Time & Exp & Log \\
  \hline
  \hline
  iris  ($3$/$4$) &  $6.5$ ($2.8$) & $0.47$s &
                $6.5$ ($2.8$) & $0.61$s &
                $\mathbf{6.4}$ ($\mathbf{3.0}$) & $8.6$s &
                $18$ & $14$\\
  glass ($6$/$9$) &  $28.4$ ($5.5$) & $1.2$s &
                $28.3$ ($5.5$) & $1.67$s &
                $\mathbf{28.2}$ ($\mathbf{5.6}$) & $7.9$s &
                $7$ & $5$\\
  usps ($10$/$256$) &  $10.4$ ($2.8$) & $2.3$s &
                $\mathbf{9.9}$ ($\mathbf{2.6}$) & $5.0$s &
                $10.5$ ($2.8$) & $26.5$s &
                $12$ & $5$ \\
  pen ($10$/$16$) &  $\mathbf{7.4}$ ($\mathbf{2.2}$) & $2.7$s &
                $\mathbf{7.4}$ ($\mathbf{2.1}$) & $5.6$s &
                $7.7$ ($2.1$) & $47.8$s &
                $18$ & $9$\\
  news ($20$/$771$) & $\mathbf{55.1}$ ($\mathbf{2.7}$) & $13.1$s &
                $55.4$ ($2.5$) & $20.3$s &
                $57.4$ ($2.8$) & $7$m$54$s &
                $36$ & $23$\\
  letter ($26$/$16$) & $25.3$ ($2.0$) & $35.7$s &
                $25.0$ ($2.1$) & $51.1$s &
                $\mathbf{24.6}$ ($\mathbf{1.9}$) & $44$m$4$s  &
                $74$ & $52$\\
  rcv1 ($38$/$1213$) & $\mathbf{33.4}$ ($\mathbf{1.7}$) &  $1$m$49$s  &
                $33.7$ ($1.6$) & $2$m$59$s  &
                $33.7$ ($1.9$) & $47$m$30$m  &
                $26$ & $16$\\
  sector ($105$/$2959$) & $\mathbf{20.1}$ ($1.2$) & $56$m$32$s &
                $20.6$ ($1.0$) & $1$h$53$m  &
                $22.3$ ($1.1$) & $16$h$37$m  &
                $17$ & $9$\\
  \hline
  \end{tabular}
  }
  \label{tab:exp_multi2}
\end{table*}

\subsection{Comparison to MultiBoost}
In this experiment, we compare our algorithm to MultiBoost, a totally corrective multi-class boosting proposed in \cite{Shen2011Direct}.
We compare both the classification accuracy and
the coefficient calculation time (training time) of our approach and MultiBoost.
For simplicity, we use decision stumps as the weak classifier.
For MultiBoost, we use the logistic loss and choose the regularization parameter from
$\{$ $10^{-8}$, $5 \times 10^{-8}$, $10^{-7}$, $5 \times 10^{-7}$, $\cdots$,
$10^{-3}$, $5 \times 10^{-3}\}$  by cross-validation.
For our algorithm, we set $\nu$ to $10^{-9}$ and $\eta$ to $0.5$.
All experiments are repeated $50$ times using the same regularization parameter.
All algorithms are implemented in MATLAB using a single processor.
The weak learner training (decision stump) is written in C and compiled as a
MATLAB MEX file.
We use MATLAB interface for L-BFGS-B \cite{CarbonettoMatlab} to solve \eqref{EQ:coeff_sw1} and \eqref{EQ:coeff_sw2}.
The maximum number of L-BFGS-B iterations is set to $100$.
The iteration stops when the projection gradient is less than $10^{-5}$ or the difference between the objective value of current iteration and previous iteration is less than $10^{-9}$.
We use the data set \emph{letter} from the UCI repository and
vary the number of classes and the number of training samples.
Experimental results are shown in Table~\ref{tab:exp_multi1} and Fig.~\ref{fig:exp_multi}.
\revised{
We observe that our approach performs comparable to MultiBoost while having a fraction of
the training time of MultiBoost.
In the next experiment, we statistically compare the proposed approach
with MultiBoost using the nonparametric Wilcoxon signed-rank
test (WSRT) \cite{Demvsar2006Statistical} on several UCI data sets.
}

In this experiment, we evaluate the proposed approach with MultiBoost
on $8$ UCI data sets.
For each data set, we randomly choose $50$ samples from each class and
randomly split the data into training and test sets at a ratio of $75$:$25$.
We repeat our experiments $50$ times.
For data sets with a large number of dimensions, we perform dimensionality reduction using PCA.
Our PCA projected data captures $90\%$ of the original data variance.
We set the number of boosting iterations to $500$.
Table~\ref{tab:exp_multi2} reports average test errors and the time it takes to compute $\bW$
of different algorithms.
\revised{
Based on our results, all methods perform very similarly.
MultiBoost has a better generalization performance than other algorithms on $3$ data sets
while \MultiEXP and \MultiLOG performs better than other algorithms
on $4$ and $2$ data sets, respectively.
We then statistically compare all three algorithms using the non-parametric
Wilcoxon signed-rank test (WSRT) \cite{Demvsar2006Statistical}.
WSRT tests the median performance difference between a pair of classifiers.
In this test, we set the significance level to be $5\%$.
The null-hypothesis declares that there is no difference between
the median performance of both algorithms at the $5\%$ significance level.
In other words, a pair of algorithms perform equally well in a statistical sense.
According to the table of exact critical values for the Wilcoxon's test,
for a confidence level of $0.05$ and $8$ data sets, the difference
between the classifiers is significant if the smaller of the
rank sums is equal or less than $3$.
For \MultiEXP and MultiBoost, the signed rank statistic result is $10.5$ and,
for \MultiLOG and MultiBoost, the result is $7$.
Since both results are not less than the critical value, WSRT indicates
a failure to reject the null hypothesis at the $5\%$ significance
level.
In other words, the test statistics suggest that both stage-wise boosting
and totally-corrective boosting perform equally well.
}
In terms of training time, both \MultiEXP and  \MultiLOG are
much faster to train compared to MultiBoost.
We have observed a significant speed-up factor (at least two orders of magnitude) depending on the complexity of the optimization problem and the number of classes.

\begin{table*}[t]
\caption{Average test errors
  and their standard deviations (shown in $\%$)
  on multi-class UCI data sets at the $1000$-th boosting iteration.
  All experiments are repeated $50$ times.
  The best average performance is shown in boldface.
  $\dag$ Results are not available due to a prohibitively amount of CPU time
  or RAM required.
  }
  \centering
  \scalebox{1}
  {
  \begin{tabular}{l|c|cccccccc}
  \hline
   \multirow{2}{*}{Dataset} & $\#$ & SAMME & Ada.ECC & Ada.MH  & Ada.MO  &  GradBoost  & MultiBoost  & \MultiEXP & \MultiLOG \\
   & classes &  \cite{Zhu2006Multi} &  \cite{Guruswami1999Multiclass} & \cite{Schapire1999Improved} & \cite{Schapire1999Improved}  & \cite{Duchi2009Boosting} & \cite{Shen2011Direct} & (ours) & (ours) \\
  \hline
  \hline
dna & 3  & $6.1$ ($0.9$)  & $7.0$ ($1.1$)  & $\mathbf{5.4}$ ($\mathbf{0.9}$)  & $6.4$ ($1.0$)  & $10.3$ ($6.4$)  & $5.7$ ($0.9$)  & $6.3$ ($0.9$)  & $6.2$ ($1.0$) \\
svmguide2 & 3  & $\mathbf{21.7}$ ($\mathbf{3.4}$)  & $22.3$ ($4.0$)  & $22.3$ ($3.4$)  & $22.8$ ($4.4$)  & $24.8$ ($3.7$)  & $22.1$ ($3.8$)  & $\mathbf{21.7}$ ($\mathbf{3.9}$)  & $22.2$ ($3.8$) \\
wine & 3  & $4.5$ ($3.6$)  & $3.6$ ($2.9$)  & $\mathbf{2.9}$ ($\mathbf{2.5}$)  & $3.1$ ($2.7$)  & $3.0$ ($3.3$)  & $3.3$ ($3.0$)  & $3.2$ ($2.9$)  & $\mathbf{2.9}$ ($\mathbf{2.5}$) \\
vehicle & 4  & $31.4$ ($2.3$)  & $22.4$ ($2.5$)  & $22.8$ ($2.3$)  & $\mathbf{21.8}$ ($\mathbf{2.7}$)  & $44.0$ ($7.7$)  & $23.6$ ($2.5$)  & $22.9$ ($2.5$)  & $22.9$ ($2.6$) \\
glass & 6  & $31.2$ ($6.2$)  & $\mathbf{27.2}$ ($\mathbf{4.7}$)  & $27.6$ ($5.7$)  & $27.4$ ($5.5$)  & $40.5$ ($7.2$)  & $28.5$ ($5.2$)  & $28.1$ ($5.8$)  & $28.2$ ($5.7$) \\
satimage & 6  & $18.0$ ($1.3$)  & $10.9$ ($0.7$)  & $10.9$ ($0.8$)  & $\mathbf{10.6}$ ($\mathbf{0.8}$)  & $19.2$ ($4.3$)  & $10.8$ ($0.8$)  & $11.1$ ($0.8$)  & $10.8$ ($0.9$) \\
svmguide4 & 6  & $24.8$ ($4.3$)  & $17.3$ ($2.7$)  & $19.3$ ($3.5$)  & $17.6$ ($3.5$)  & $20.1$ ($5.1$)  & $17.3$ ($3.1$)  & $\mathbf{16.6}$ ($\mathbf{2.7}$)  & $\mathbf{16.6}$ ($\mathbf{2.8}$) \\
segment & 7  & $10.3$ ($3.4$)  & $\mathbf{2.1}$ ($\mathbf{0.6}$)  & $2.9$ ($0.8$)  & $2.4$ ($0.8$)  & $2.8$ ($0.7$)  & $2.5$ ($0.9$)  & $2.3$ ($0.8$)  & $2.2$ ($0.7$) \\
usps & 10  & $16.5$ ($2.1$)  & $8.9$ ($1.6$)  & $8.9$ ($1.5$)  & $\mathbf{8.5}$ ($\mathbf{1.5}$)  & $11.4$ ($1.7$)  & $9.0$ ($1.6$)  & $9.0$ ($1.4$)  & $8.7$ ($1.6$) \\
pendigits & 10  & $15.8$ ($2.4$)  & $\mathbf{4.5}$ ($\mathbf{1.1}$)  & $4.9$ ($1.3$)  & $5.5$ ($1.5$)  & $5.2$ ($1.4$)  & $4.8$ ($1.3$)  & $4.9$ ($1.4$)  & $4.9$ ($1.3$) \\
vowel & 11  & $40.5$ ($3.0$)  & $8.1$ ($1.6$)  & $10.5$ ($1.9$)  & $12.1$ ($2.0$)  & $38.7$ ($5.1$)  & $7.5$ ($1.6$)  & $\mathbf{6.5}$ ($\mathbf{1.6}$)  & $\mathbf{6.5}$ ($\mathbf{1.5}$) \\
news & 20  & $72.5$ ($3.2$)  & $63.1$ ($3.0$)  & $63.5$ ($5.9$)  & $ \dag $  & $66.9$ ($2.6$)  & $\mathbf{52.8}$ ($\mathbf{2.5}$)  & $55.0$ ($3.5$)  & $54.5$ ($3.2$) \\
letter & 26  & $52.6$ ($3.4$)  & $28.2$ ($2.0$)  & $26.5$ ($2.1$)  & $ \dag $  & $44.0$ ($3.8$)  & $24.6$ ($2.6$)  & $24.3$ ($2.1$)  & $\mathbf{24.2}$ ($\mathbf{2.0}$) \\
isolet & 26  & $ \dag $  & $9.1$ ($0.9$)  & $7.5$ ($0.9$)  & $ \dag $  & $12.7$ ($2.3$)  & $6.3$ ($0.8$)  & $6.9$ ($0.9$)  & $\mathbf{6.1}$ ($\mathbf{0.8}$) \\
  \hline
  \end{tabular}
  }
  \label{tab:exp_UCI}
\end{table*}

\begin{figure*}[t]
    \begin{center}
        \includegraphics[width=0.235\textwidth,clip]{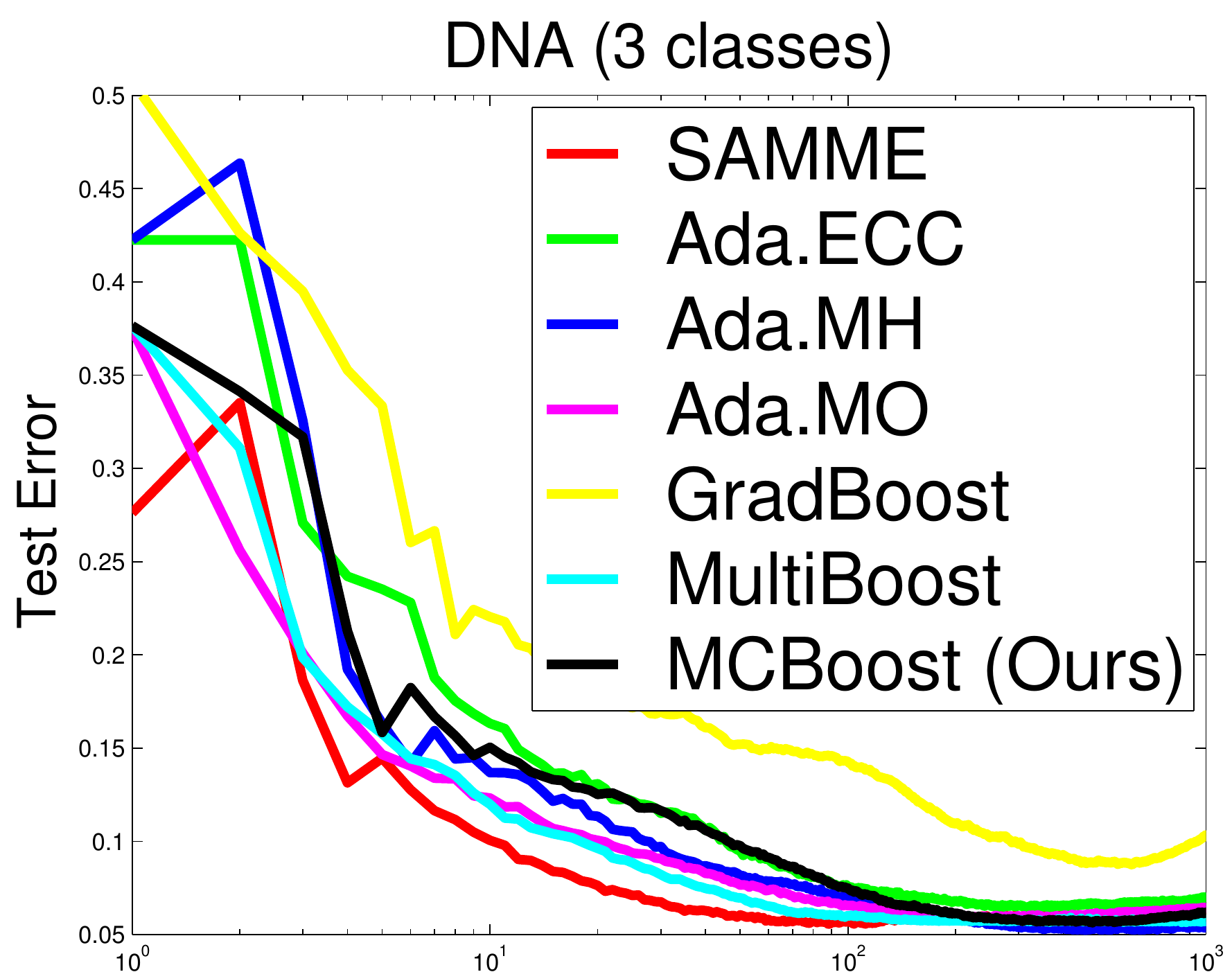}
        \includegraphics[width=0.22\textwidth,clip]{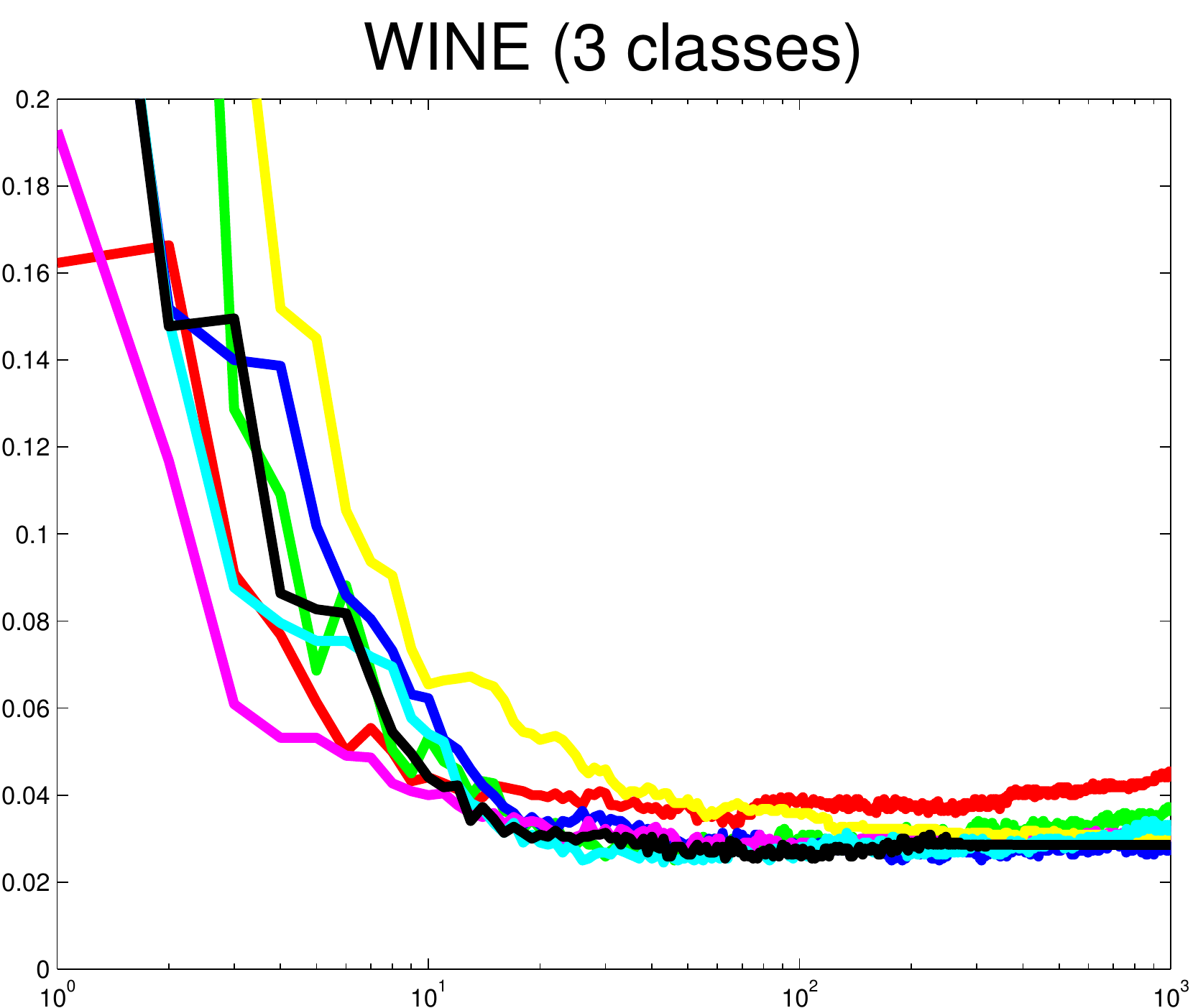}
        \includegraphics[width=0.22\textwidth,clip]{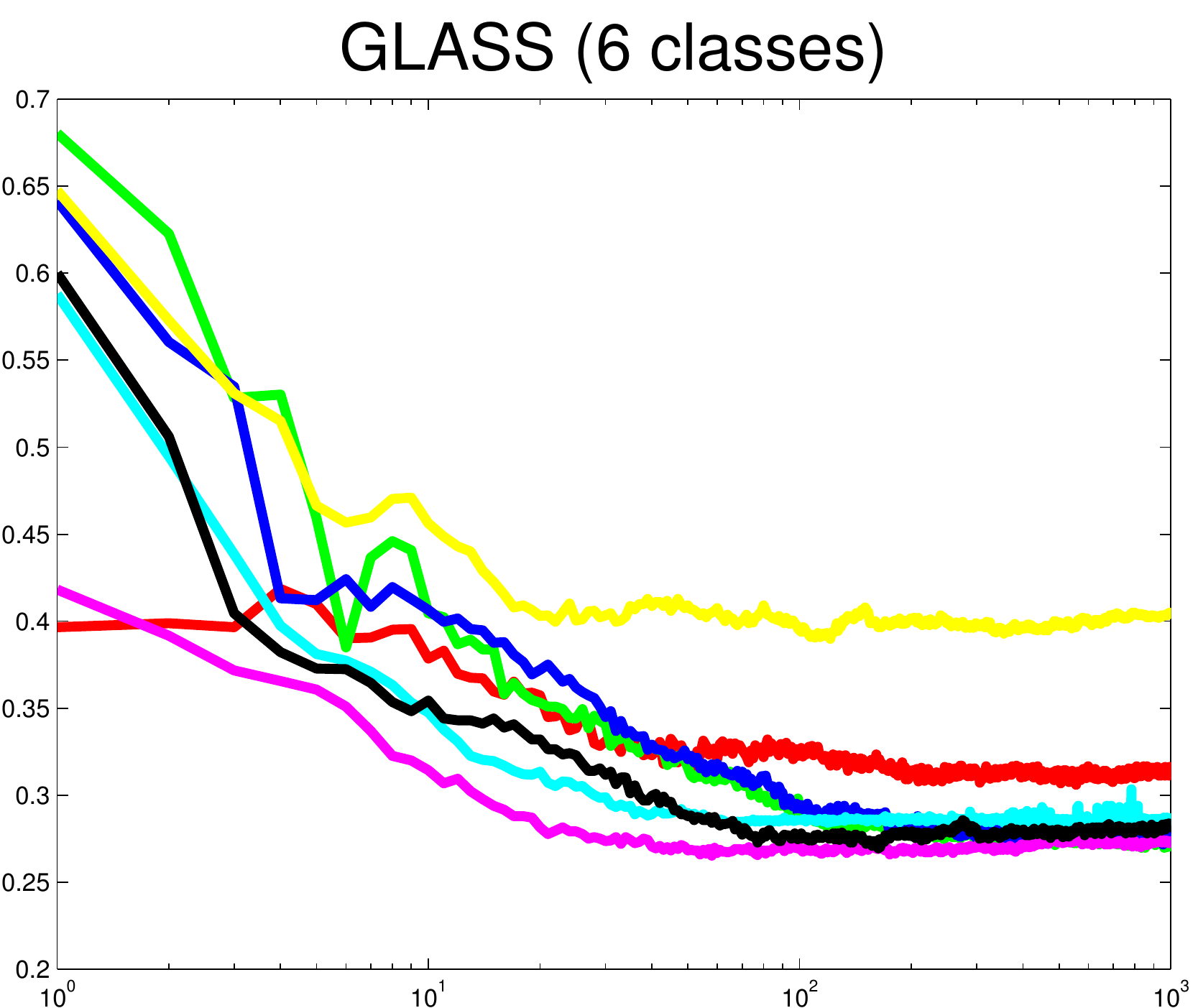}
        \includegraphics[width=0.22\textwidth,clip]{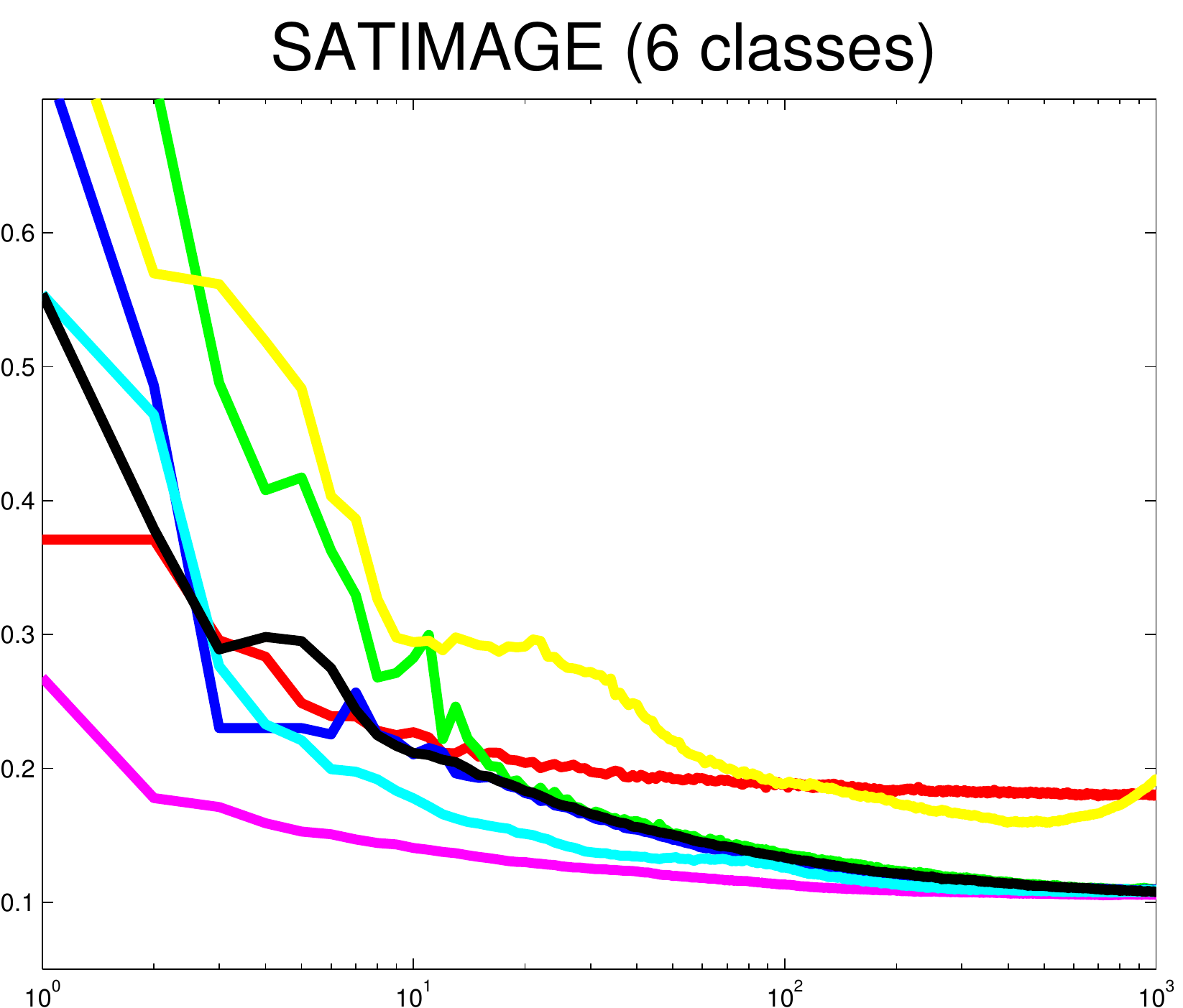}
        \includegraphics[width=0.235\textwidth,clip]{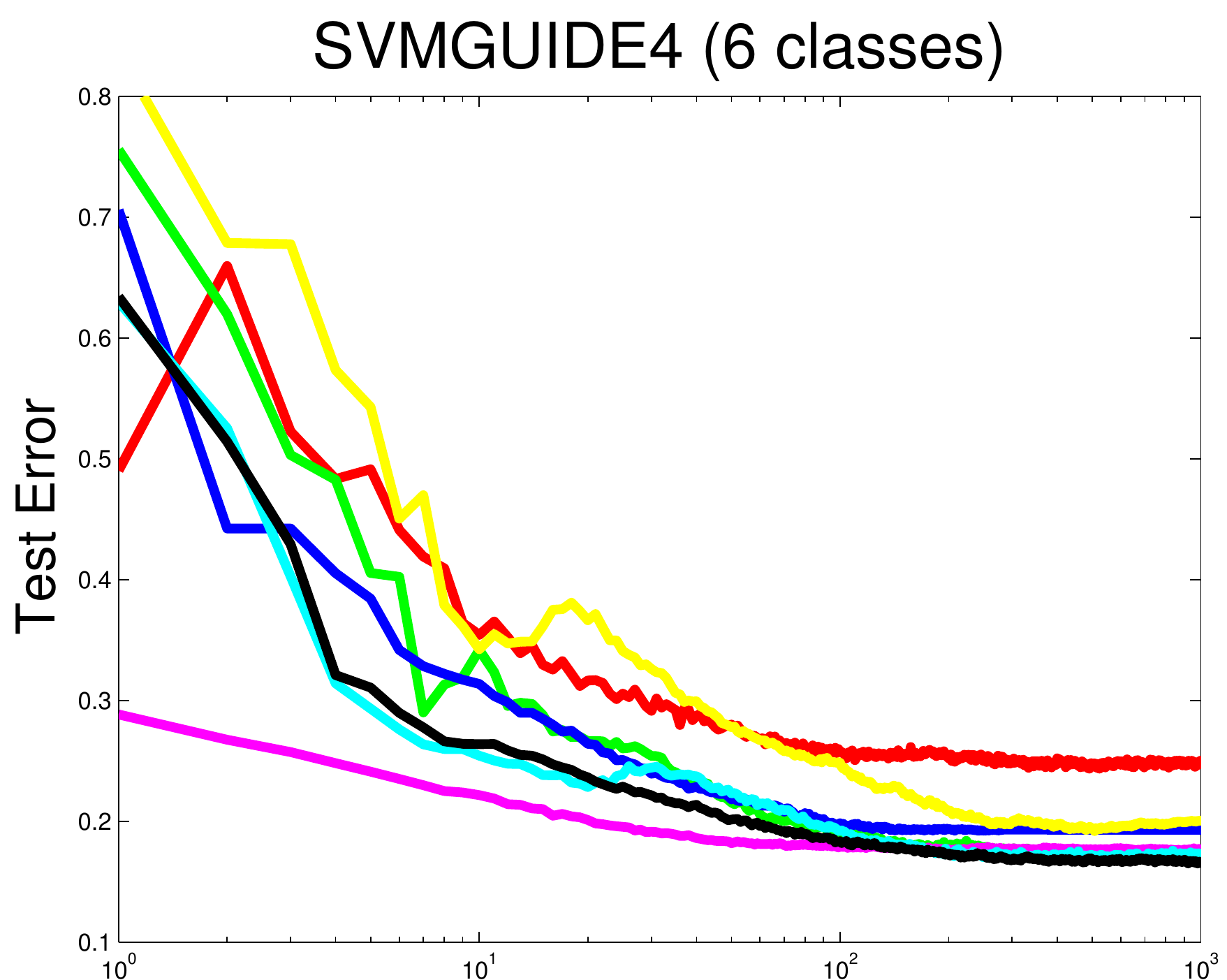}
        \includegraphics[width=0.22\textwidth,clip]{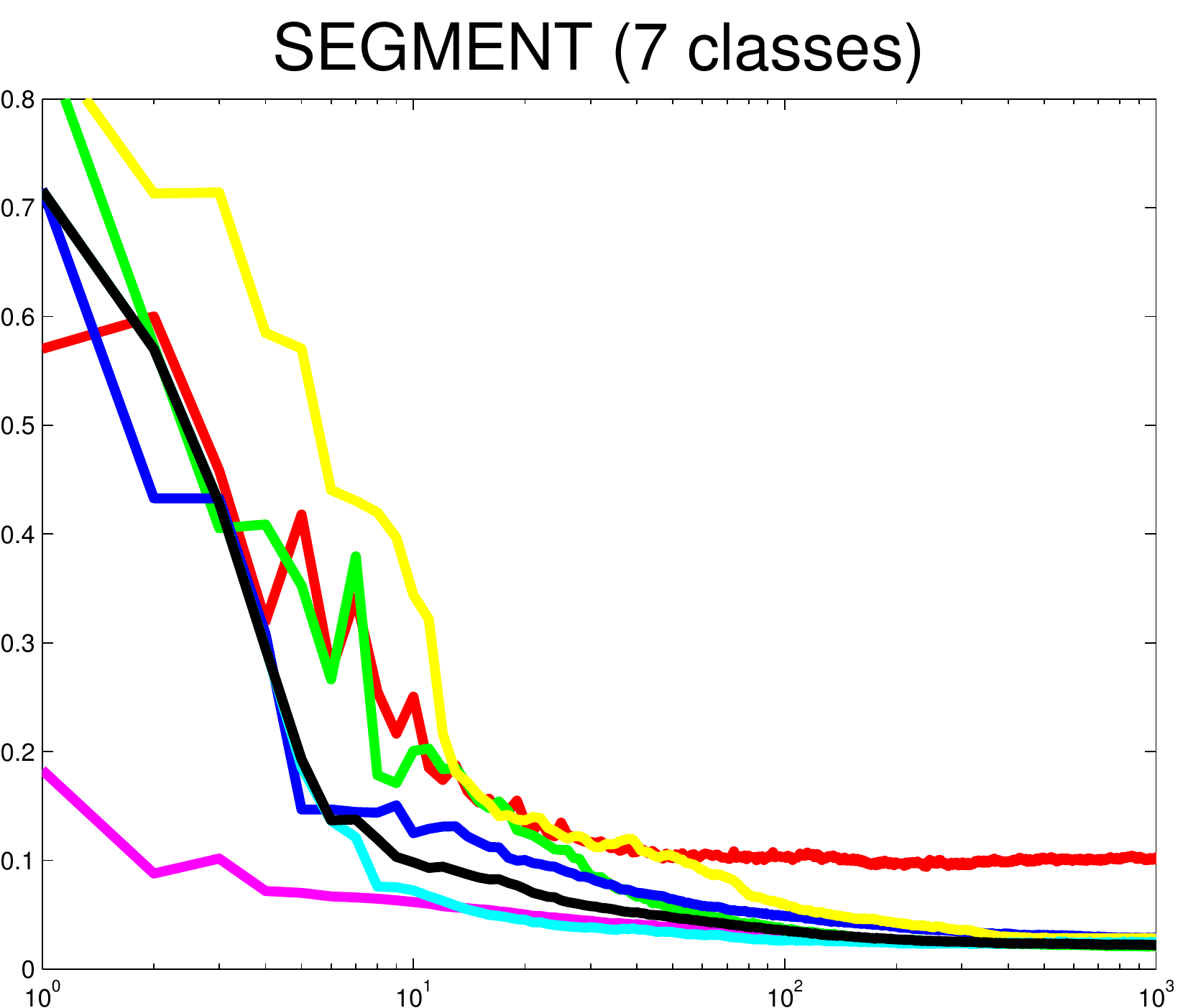}
        \includegraphics[width=0.22\textwidth,clip]{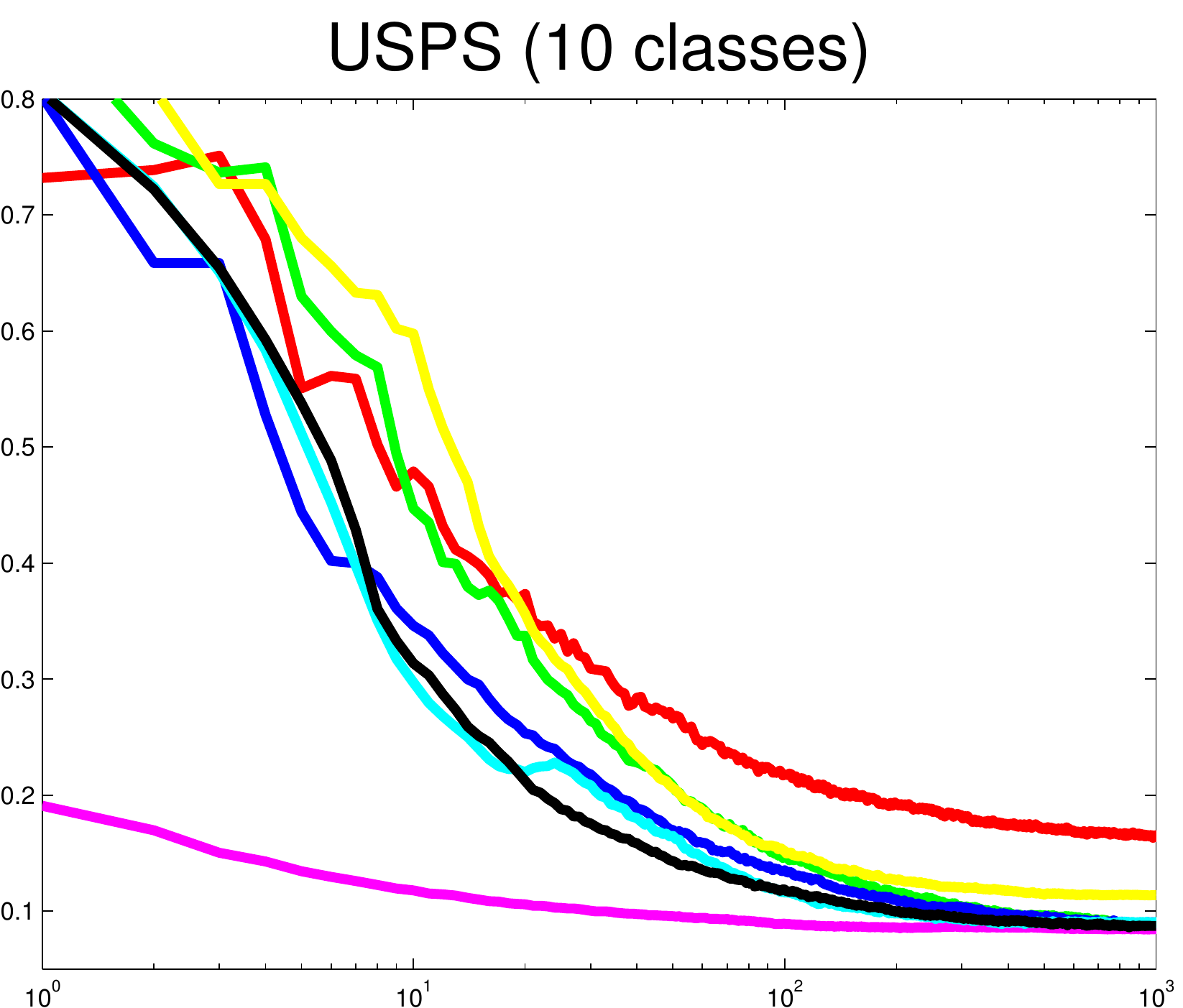}
        \includegraphics[width=0.22\textwidth,clip]{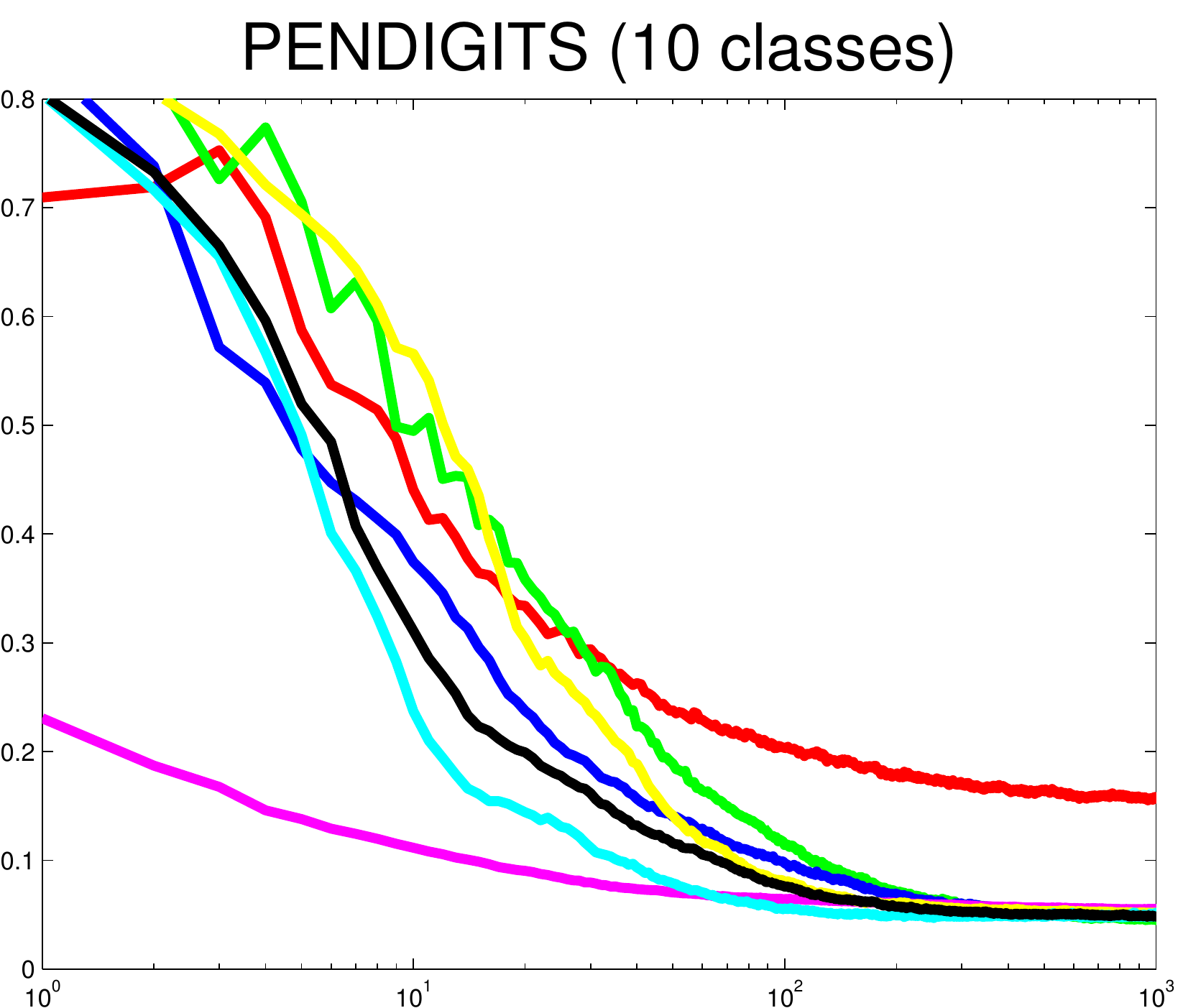}
        \includegraphics[width=0.22\textwidth,clip]{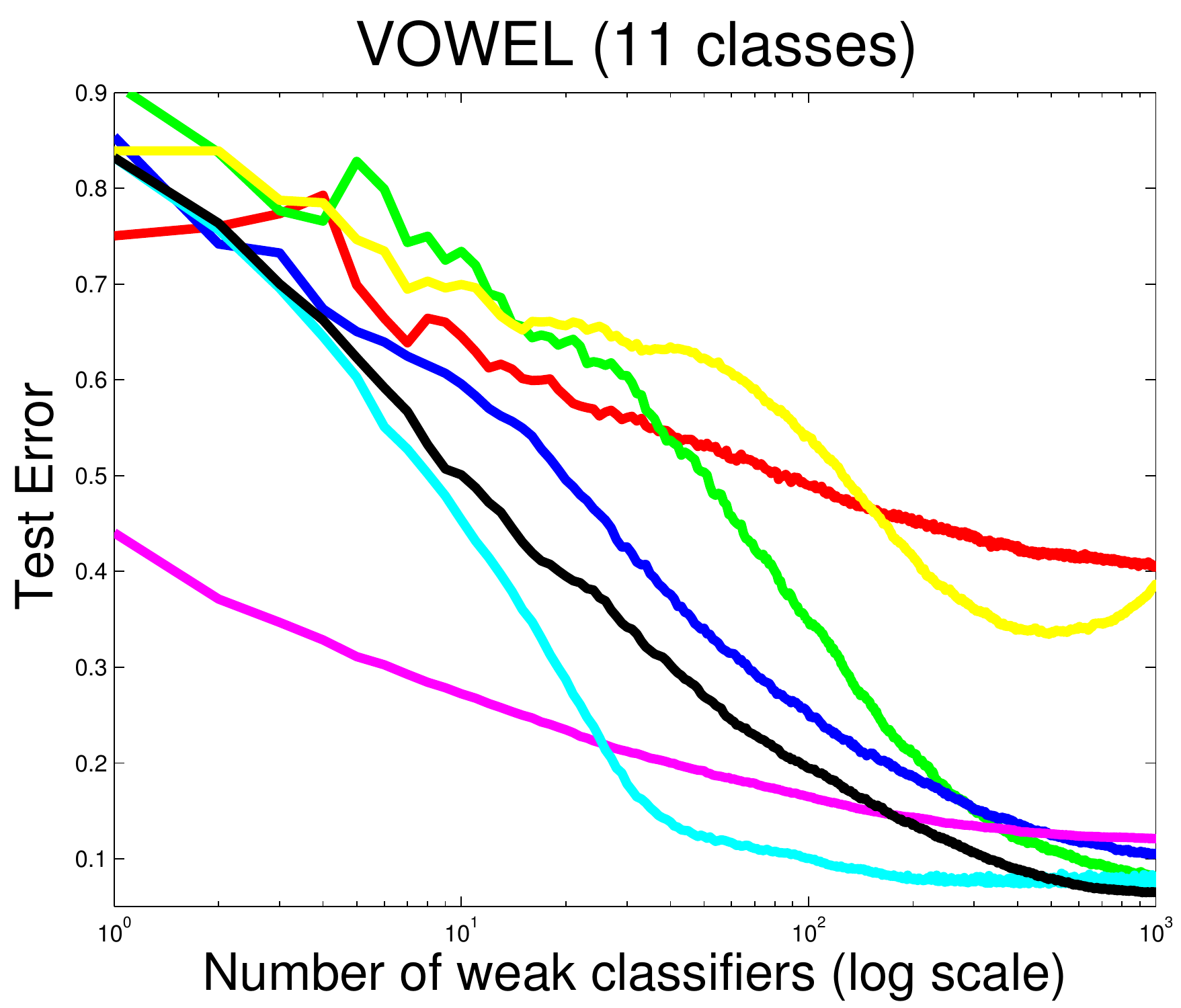}
        \includegraphics[width=0.22\textwidth,clip]{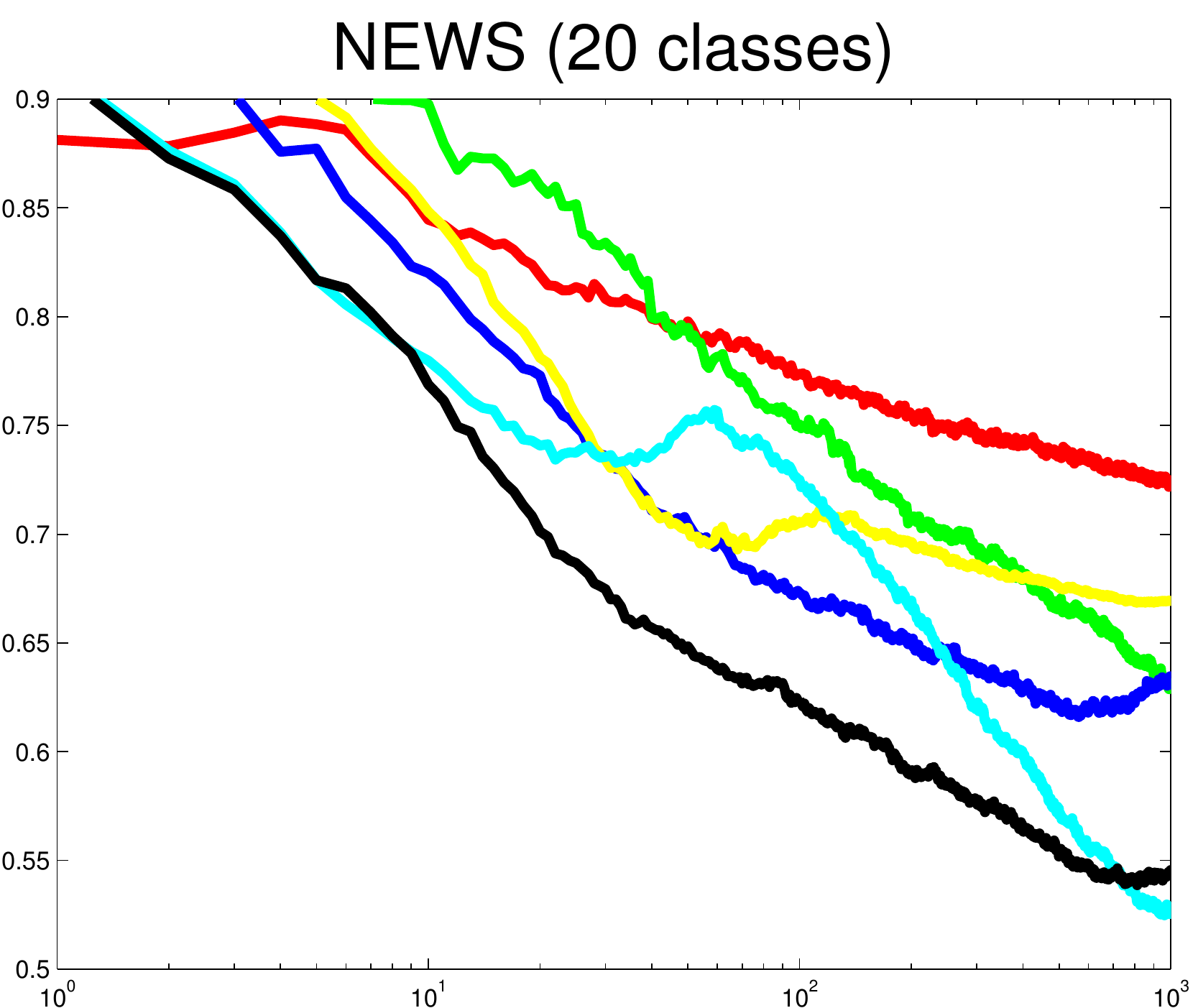}
        \includegraphics[width=0.22\textwidth,clip]{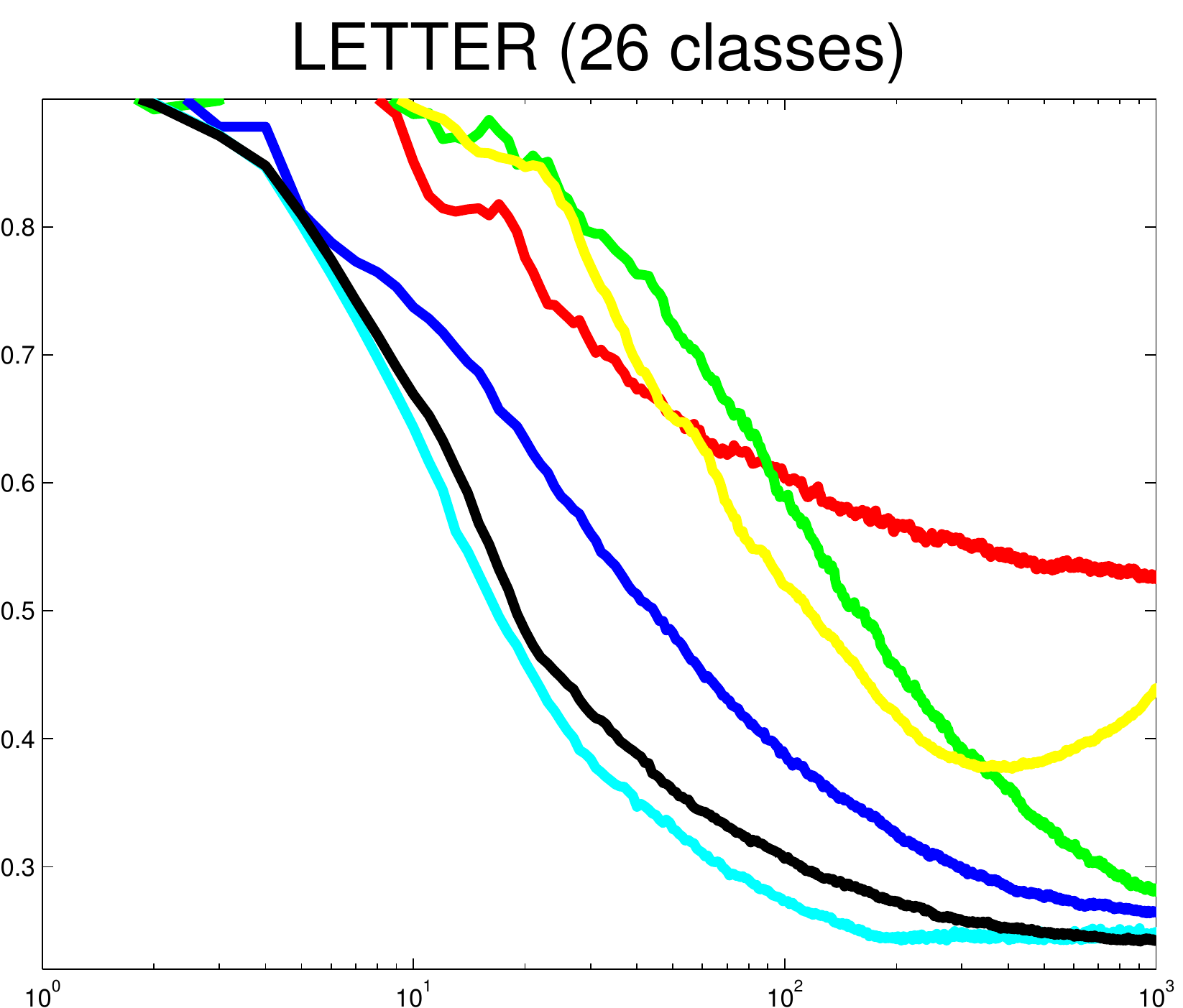}
        \includegraphics[width=0.22\textwidth,clip]{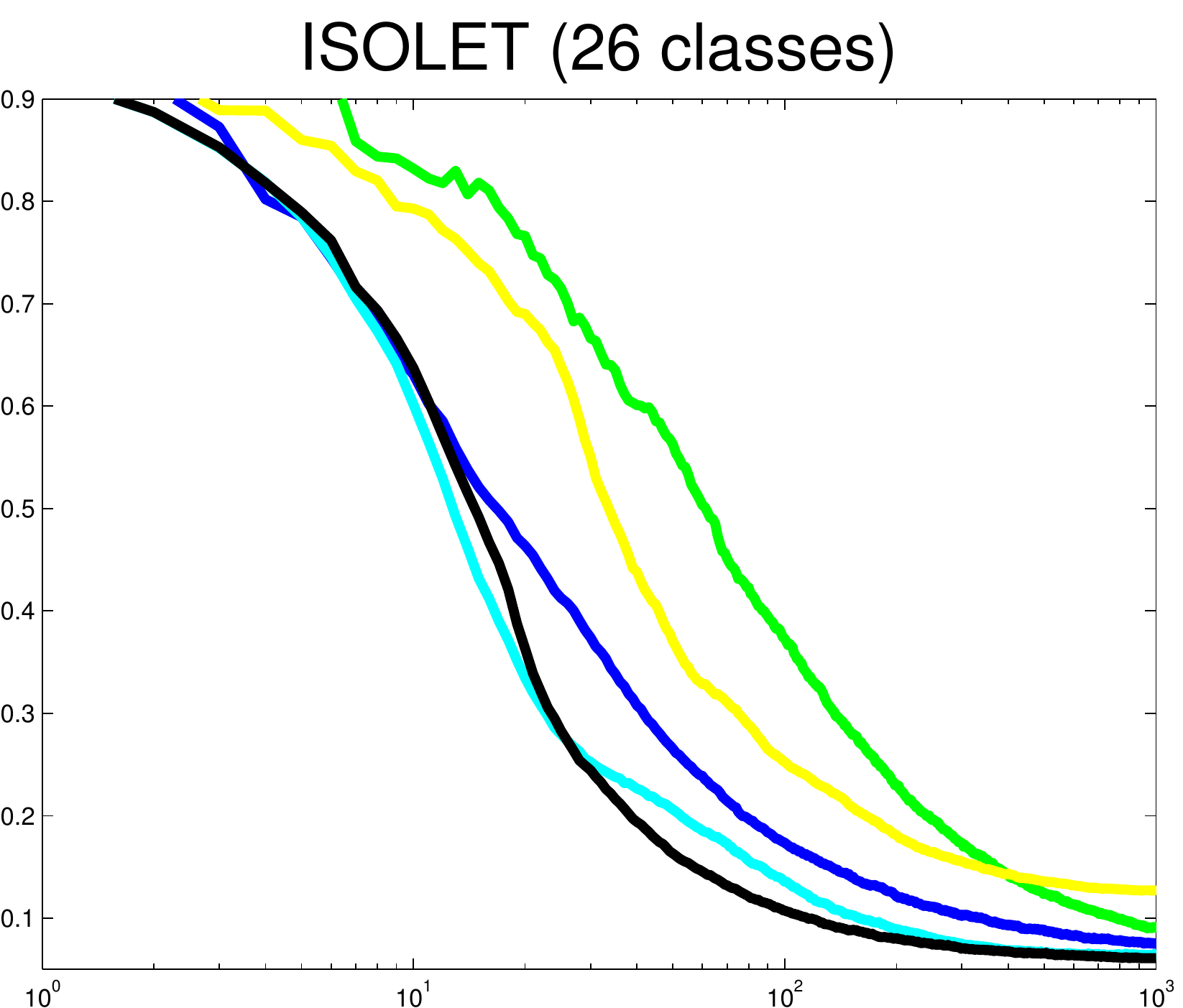}
    \end{center}
    \vspace{-.2cm}
    \caption{
    Average test error versus number of weak classifiers (logarithmic scale)
    on multi-class UCI data sets.
    The vertical axis denotes the averaged test error rate and the horizontal axis denotes the number of weak classifiers.
    On UCI benchmark data sets, our approach performs comparable to other algorithms.
    }
    \label{fig:exp_UCI}
\end{figure*}

\begin{table*}
\caption{Comparison of multi-class boosting with their evaluation functions and
average evaluation time per test instance on pendigits data set ($10$
classes).
  }
  \centering
  {
  \begin{tabular}{lllc}
  \hline
  Algorithm & Evaluation function & Coefficients & Test time (msecs) \\
  \hline
  \hline
  Coding based (single-label)  & \multirow{2}{*}{$F(\bx) = \argmax_{r=1,\cdots,k} \, \sum_{j=1}^{n} w_j \hbar_j(\bx) m_{jr}$}
        & $\bM \in \Real^{n \times k}$, $\bw \in \Real^n$ & $0.11$ \\
  \eg, Ada.MH \cite{Schapire1999Improved}, Ada.ECC \cite{Guruswami1999Multiclass} & & & \\
  \hline
  Coding based (multi-label)  & \multirow{2}{*}{$F(\bx) = \argmax_{r=1,\cdots,k} \, \sum_{j^{\prime} \in \mathcal{Y}^{\prime}}
        m_{j^{\prime}r} \sum_{j=1}^{n} w_j \hbar_{(j^\prime,j)} (\bx) $}
        & $\bM \in \Real^{k^\prime \times k}$, $\bw \in \Real^n$,  & $7.90$ \\
  \eg, AdaBoost.MO \cite{Schapire1999Improved} & & $k^\prime = |\mathcal{Y}^{\prime}| = 2^{k-1}-1$ & \\
  \hline
  A matrix of coefficient \eg, \MultiEXPshort, & \multirow{2}{*}{$F(\bx) =  \argmax_{r = 1,\cdots,k} {\textstyle \sum_{j=1}^n }
        \hbar_j(\bx) w_{jr},$}
        & $\bW \in \Real^{n \times k}$ & $0.06$ \\
   \MultiLOGshort, MultiBoost \cite{Shen2011Direct}, GradBoost \cite{Duchi2009Boosting} & & & \\
  \hline
  \end{tabular}
  }
  \label{tab:evalfn}
\end{table*}

\subsection{Multi-class boostings on UCI data sets}
Next we compare our approaches against some well known multi-class boosting algorithms:
SAMME \cite{Zhu2006Multi},
AdaBoost.MH \cite{Schapire1999Improved},
AdaBoost.ECC \cite{Guruswami1999Multiclass},
AdaBoost.MO \cite{Schapire1999Improved},
GradBoost \cite{Duchi2009Boosting} and
MultiBoost \cite{Shen2011Direct}.
For AdaBoost.ECC, we perform binary partition using the random-half method \cite{Li2006Multiclass}.
For GradBoost, we implement $\ell_1$/$\ell_2$-regularized multi-class boosting and choose the regularization parameter from
$\{$ $5 \times 10^{-7}$, $10^{-6}$, $5 \times 10^{-6}$, $\cdots$,
$5 \times 10^{-2}$, $10^{-1}\}$
All experiment are repeated $50$ times.
The maximum number of boosting iterations is set to $1000$.
Average test errors of different algorithms and their standard deviations (shown in $\%$)
are reported in Table~\ref{tab:exp_UCI}.
On UCI data sets, we observe that the performance of most methods are comparable.
However, \MultiLOG has a better generalization performance than other
multi-class boosting algorithms on $5$ out of $14$ data sets evaluated.
In addition, directly maximizing the multi-class margin (as in MCBoost
and MultiBoost) often leads to better generalization performance in our experiments
(especially on the data set in which the number of classes
is larger than $10$).
Note that similar findings have also been reported in \cite{Daniely2012Multiclass} where
the authors theoretically compare different multi-class classification
algorithms.
They concluded that learning a matrix of coefficients, \ie,
the multi-class formulation of \cite{Crammer2002Algorithmic}, should be preferred to
other multi-class learning methods.

We also plot average test errors versus number of weak classifiers
on a logarithmic scale in Fig.~\ref{fig:exp_UCI}.
From the figure, AdaBoost.MO has the fastest convergence rate
followed by MultiBoost and our proposed approach.
AdaBoost.MO has the fastest convergence rate since it
trains $2^{k−1} − 1$ weak classifiers
at each iteration, while other multi-class algorithms train $1$ weak
classifier at each iteration.
For example, on USPS digit data
sets, the AdaBoost.MO model would have a total of $511,000$
weak classifiers ($1000$ boosting iteration) while all other multi-class
classifiers would only have $1000$ weak classifiers.
A comparison of evaluation functions of different multi-class classifiers
is shown in Table~\ref{tab:evalfn}.
We also illustrate the evaluation time of different functions on
$10$ classes pendigits data set.
AdaBoost.MO is much slower than other algorithms during test time.
From Fig.~\ref{fig:exp_UCI}, the convergence rate of MultiBoost is slightly faster than
our approach since MultiBoost adjusts classifier weights ($\bW$)
in each iteration.
Our algorithm does not have this property and converge slower than
MultiBoost.
However, both algorithms achieve similar classification accuracy
when converged.
Note that our experiments are performed on an Intel core i-$7$ CPU $930$ with $12$ GB memory.

\subsection{MNIST handwritten digits}
Next we evaluate our approach on well known handwritten digit data sets.
We first resize the original image to a resolution of $28 \times 28$ pixels and apply a de-skew pre-processing.
We then apply a spatial pyramid and extract $3$ levels of HOG features with $50\%$ block overlap.
The block size in each level is $4 \times 4$, $7 \times 7$ and $14 \times 14$ pixels, respectively.
Extracted HOG features from all levels are concatenated.
In total, there are $2,172$ HOG features.
We train our classifiers using $60,000$ training samples and test it on the original test sets of $10,000$ samples.
We train $1000$ boosting iterations and the results are briefly summarized in Table~\ref{tab:ABCDETC_error}.
Our algorithm performs best compared to other evaluated algorithms.

\begin{table}
  \caption{Test errors rate (\%) on the MNIST data set
  }
  \centering
  \scalebox{1}
  {
  \begin{tabular}{l|c|c}
  \hline
   $$ & $\#$ features & Accuracy ($\%$) \\
   \hline
   \hline
   AdaBoost.MH \cite{Schapire1999Improved} & $1000$  & $1.29$ \\
   AdaBoost.ECC \cite{Guruswami1999Multiclass} & $1000$ & $1.24$  \\
   \MultiEXPshort (ours) & $1000$ & $1.14$ \\
   \MultiLOGshort (ours) & $1000$ & $1.03$ \\
  \hline
  \end{tabular}
  }
  \label{tab:ABCDETC_error}
\end{table}

\subsection{Scene recognition}
In the next experiment, we compare our approach on the
$15$-scene data set used in
\cite{Lazebnik2006Beyond}.
The set consists of $9$ outdoor scenes and 6 indoor
scenes.
There are $4,485$ images in total.
For each run, we randomly split the data into a training set and a test set based on published protocols.
This is repeated $5$ times and the average accuracy is reported.
In each train/test split, a visual codebook is generated.
Both training and test images are then transformed into histograms of code words.
\revised{
We use CENTRIST as our feature descriptors \cite{Wu2011CENTRIST}.
We build $200$ visual code words using the histogram intersection kernel (HIK)
on the training set.
We represent each image in a spatial hierarchy manner \cite{Bosch2008Scene}.
Each image is divided into $3$ levels.
The first level divides the image into $4 \times 4$ blocks and
$3 \times 3$ blocks.
The second level divides the image into $2 \times 2$ blocks and $1 \times 1$ block and
the last level is the image itself.
The total number of sub-windows in all $3$ levels are $31$.
Note that the image is resized between different levels so that all blocks contain
the same number of pixels.
An image is then represented by the concatenation of histograms of code words
from all $31$ sub-windows.
Hence, in total there are $200 \times 31$ dimensional histogram \cite{Wu2011CENTRIST}.
To learn a weak classifier, we select a subset of discriminative CENTRIST features.
For example, a decision stump would select the most discriminative CENTRIST feature
while a decision tree would select several CENTRIST features depending on
the depth of the decision tree.
}

\begin{figure}[t!]
  \centering
        \includegraphics[width=0.35\textwidth,clip]{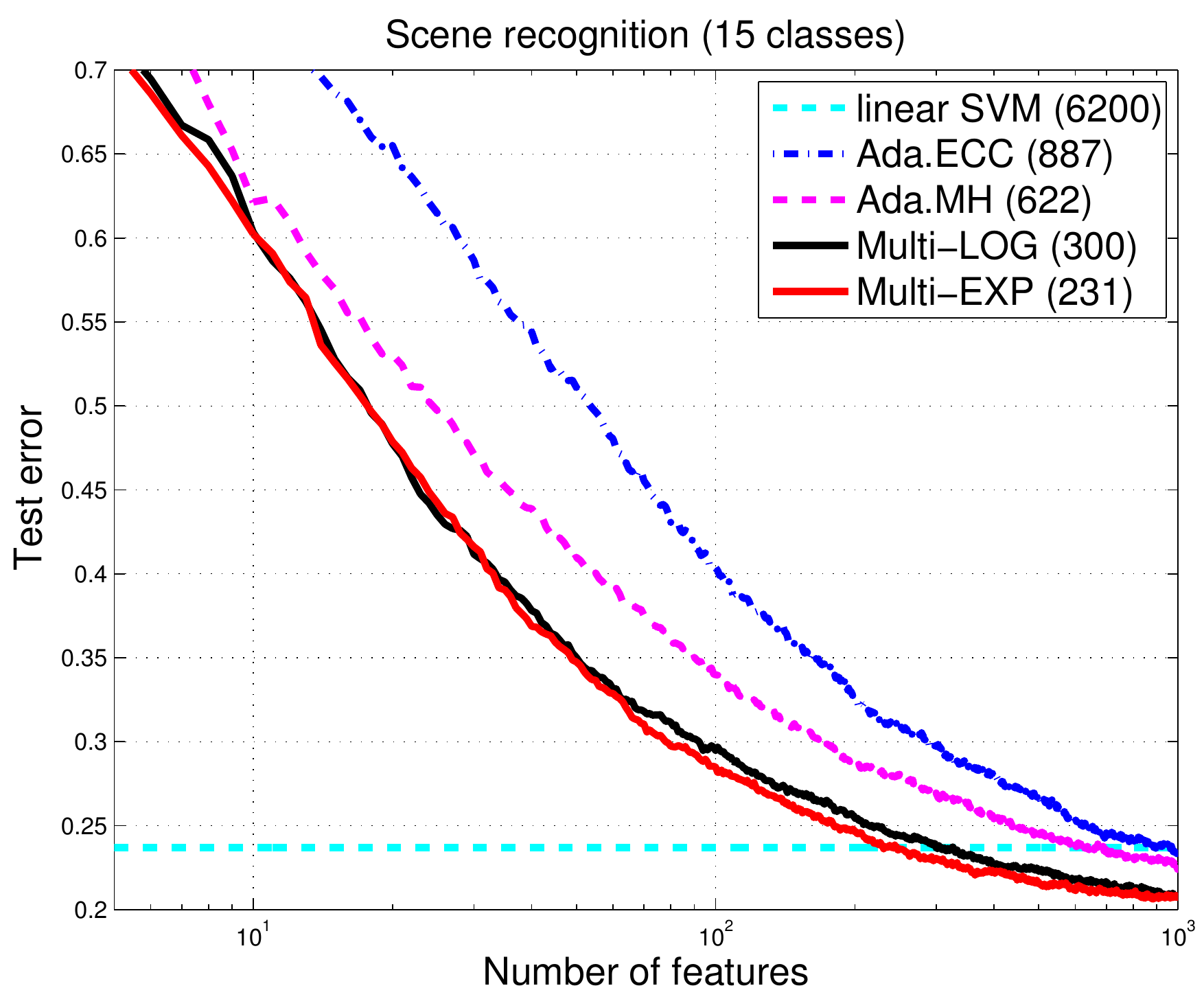}
    \caption{
    Performance of different classifiers on the scene recognition data set.
    We also report the number of features required to achieve similar results
    to linear multi-class SVM.
    Both of our methods outperform other evaluated boosting algorithms.
    }
    \label{fig:scene}
  \end{figure}

\secondrev{
Fig.~\ref{fig:scene} shows the average classification errors.
Based on our experimental results, our approaches have the fastest convergence rate
and the lowest test error compared to other algorithms evaluated.
We also apply a multi-class SVM to the above data set using the
LIBSVM package \cite{Chang2011LIBSVM} and
report the recognition results in Table~\ref{tab:scene}.
We train SVM using CENTRIST features ($6200$ dimensions)
and PCA-CENTRIST features ($1000$ dimensions\footnote{
We reduce the feature dimension by projecting CENTRIST
features to the new subspace using eigenvectors
of $1000$ largest magnitude eigenvalues.}).
For CENTRIST, we train the multi-class SVM using linear and
histogram intersection kernel (HIK).
For PCA-CENTRIST, we train the multi-class SVM using linear
and radial basis function kernel (RBF).
We set the maximum number of boosting iterations to $1000$ and $6200$
for a fair comparison between SVM and multi-class boosting.
At $1000$ features, our approach with the exponential loss performs best.
Interestingly, RBF SVM only performs slightly better than linear SVM on PCA-CENTRIST features.

We then increase the number of boosting iterations to $6200$.
We observe that our approach performs comparable to
MultiBoost while having a fraction of the training time of MultiBoost.
We also note that AdaBoost.ECC has a slightly better generalization performance
than our approach at $6200$ iterations.
We suspect that this is due to its slow convergence rate which helps avoid
the problem of over-fitting.
We observe that Linear SVM with $6200$ features performs slightly better than
Linear SVM with $1000$ features.
This may be due to the fact that we have not fine tuned the cross
validation parameter for PCA-CENTRIST features.
From Table~\ref{tab:scene}, HIK SVM with $6200$ features
achieves the highest classification accuracy.
However, we observe that HIK SVM has the largest computational complexity during evaluation time.
Based on the efficient SVM implemenation of \cite{Chang2011LIBSVM} and \cite{Wu2009Beyond},
the average evaluation time of linear and non-linear SVM per test instance on
the scene-$15$ data sets is $16$ milliseconds and $19$ milliseconds, respectively.
In contrast, the average evaluation time per test instance of
our approach is only $0.08$ milliseconds
(a speed up by more than two orders of magnitude).
It is important to note that we use a fast evaluation of
HIK SVM (using a pre-compuated table) \cite{Wu2009Beyond} in our experiment.
For other non-linear families of kernels,
the computational complexity of the non-linear SVM becomes much more expensive,
\ie, the cost to compute the kernel function values for RBF is
$\bigO(d \cdot | \pi |))$ summations
and $\bigO( | \pi | )$ exponential operations
where $d$ is the number of histogram bins and
$| \pi |$ is the cardinality of the set of support vectors.
As a large number of support vectors are usually generated during training,
this leads to a high computational complexity during evalution time.
Another drawback of HIK SVM is that
the classifier is only applicable to histogram features where
the value is a non-negative real number.
The classifier cannot be direclty applied to many machine learning and computer vision data sets.
In summary, our approach achieves the best trade-off in terms of speed and accuracy.
This saving in computation time during evaluation is particularly {\em important
for many real-time applications} in which fewer number of features are preferred.
}

\begin{table}[t!]
  \caption{Recognition rate of various algorithms on Scene-$15$ data sets.
  All experiments are repeated $5$ times.
  The average accuracy mean and standard deviation (in percentage) are
  reported.
  }
  \centering
  {
  \begin{tabular}{l|c|c}
  \hline
   $$ & $\#$ features & Accuracy ($\%$) \\
   \hline
   \hline
   Linear SVM & $1000$ & $74.5$ ($0.7$) \\
   Non-linear (RBF) SVM & $1000$ & $74.8$ ($0.6$) \\
   AdaBoost.SIP \cite{Zhang2009Finding} & $1000$ &  $75.7$ ($0.1$) \\
   AdaBoost.ECC \cite{Guruswami1999Multiclass} & $1000$ & $76.5$ ($0.7$) \\
   AdaBoost.MH \cite{Schapire1999Improved} & $1000$  & $77.6$ ($0.6$) \\
   MultiBoost \cite{Shen2011Direct} & $1000$ & $79.1$ ($0.2$) \\
   \MultiLOGshort (ours) & $1000$ & $79.1$ ($0.6$) \\
   \MultiEXPshort (ours) & $1000$ & $\mathbf{79.3}$ ($\mathbf{0.5}$) \\
   \hline
   Linear SVM & $6200$ & $76.3$ ($0.9$) \\
   AdaBoost.MH \cite{Schapire1999Improved} & $6200$  & $80.0$ ($0.3$) \\
   \MultiLOGshort (ours) & $6200$ & $80.1$ ($0.6$) \\
   \MultiEXPshort (ours) & $6200$ & $80.3$ ($0.5$) \\
   MultiBoost \cite{Shen2011Direct} & $6200$ & $80.5$ ($0.5$) \\
   AdaBoost.ECC \cite{Guruswami1999Multiclass} & $6200$ & $80.8$ ($0.2$) \\
   Non-linear (HIK) SVM  & $6200$ & $\mathbf{81.5}$ ($\mathbf{0.6}$) \\
  \hline
  \end{tabular}
  }
  \label{tab:scene}
\end{table}

\section{Conclusion}
\label{sec:con}

In this paper, we focus primarily on the direct formulation of multi-class boosting methods.
Unlike many existing multi-class boosting algorithms, which rely on error correcting code, we directly maximize the multi-class margin in a stage-wise manner.
Reformulating the problem this way enables us to speed up the training time while maintaining the high classification performance.
Various multi-class boosting algorithms are thoroughly evaluated on a multitude of multi-class data sets.
Empirical results reveal that the new approach can speed up the classifier training time by more than two orders of magnitude without sacrificing detection accuracy.
On visual object classification problems, e.g., MNIST and Scene-$15$, it is beneficial to apply our approach due to its fast convergence and better accuracy has been observed compared to coding-based approach.
Directions for possible future works include applying the proposed approach
to other object detection applications, particularly those requiring real-time
performance.
Scalability of the proposed approach with respect to the number
of classes could also be explored.

\bibliographystyle{ieee}
\bibliography{tnn_boosting}

\end{document}

%% file: alg1.tex
\SetKwInput{KwInit}{Initilaize}

\begin{algorithm}[t]
\caption{Stage-wise based multi-class boosting
}
%
%
\footnotesize{
   \KwIn{
     \\1)    A set of examples $\{\bx_i,y_i\}$, $i=1 \cdots m$;
     \\2)    The maximum number of weak classifiers, $n$;
   }

   \KwOut{
      A multi-class classifier $F(\bx) = \argmax_{r = 1,\cdots,k} \sum_{j=1}^{n} \hbar_j( \bx) w_{jr} $
}

\KwInit {
   \\1)      $j \leftarrow 0$;

   \\2)      Initialize sample weights, $u_{ir} = \frac{1}{mk}$;
}

\While{ $j < n$ }
{
  \cone\ Train a weak learner, $\hbar_j(\cdot)$, by solving the subproblem \eqref{EQ:weak2};
  \\ \ctwo\ If the stopping criterion,
  ${\textstyle \sum_{i}} \left[ \delta_{r, \yi}
            \left( {\textstyle \sum_{l=1}^k} u_{il} \right) - u_{ir} \right] \hbar_j(\bxi) \leq \nu + \epsilon, \forany r$,
   has been met, we exit the loop;
  \\ \cthree\ Add the best weak learner, $h_j(\cdot)$, into the current set;
  \\ \cfour\ Solve the primal problem: \eqref{EQ:coeff_sw1} for exponential loss or \eqref{EQ:coeff_sw2} for logistic loss;
  \\ \cfive\ Update sample weights (dual variables), \eqref{EQ:KKT1} or \eqref{EQ:KKT2};
  \\ \csix\ $j \leftarrow j + 1$;
}
} %
%
\label{ALG:alg1}
\end{algorithm}